\documentclass[10pt,final,journal]{IEEEtran}
\pdfoutput=1
\usepackage{cite,color,amsmath,graphicx,subfigure,eucal,bm,epstopdf}
\usepackage{amssymb}
\usepackage{hyperref}
\usepackage{multirow,multicol,threeparttable,booktabs}
\usepackage{makecell}
\usepackage[table]{xcolor}
\usepackage{algorithm}

\usepackage{algorithmicx}
\usepackage{algpseudocode}
\algdef{SE}[DOWHILE]{Do}{doWhile}{\algorithmicdo}[1]{\algorithmicwhile\ #1}%

\usepackage{pgfplots,tikz}
\pgfplotsset{compat=1.15}
\usepackage{tikz}
\usetikzlibrary{shapes,arrows,spy,positioning,patterns,shapes.geometric}
\usetikzlibrary{external}
\newtheorem{theorem}{Theorem}
\newtheorem{lemma}{Lemma}

\newtheorem{conjecture}{Conjecture}

\newcommand{\Z}{\mathbb{Z}}
\newcommand{\prob}{PMP-D}

\definecolor{newhskkcolor}{rgb}{0, 0.4470, 0.7410}
\definecolor{darkgreen}{rgb}{0.0, 0.5, 0.0}

\title{\LARGE \bf
Bounds on Optimal Revisit Times in Persistent Monitoring Missions with a Distinct \& Remote Service Station}
\author{
S. K. K. Hari,~\IEEEmembership{Member,~IEEE}, S. Rathinam,~\IEEEmembership{Senior Member,~IEEE}, S. Darbha,~\IEEEmembership{Fellow,~IEEE}, K. Kalyanam,~\IEEEmembership{Senior Member,~IEEE}, S. G. Manyam and D. Casbeer,~\IEEEmembership{Senior Member,~IEEE}%
\thanks{S. K. K. Hari is with the Applied Mathematics and Plasma Physics Division, Los Alamos National Laboratory, NM, 87544 (email: hskkanth@gmail.com).}
\thanks{S. Rathinam and S. Darbha are with the Department of Mechanical Engineering, Texas A\&M University, College Station, TX 77843.}%
\thanks{K. Kalyanam is with the NASA Ames Research Center, Moffett Field, CA 94035.}%
\thanks{S. G. Manyam is with the Infoscitex corporation, Dayton, OH 45431.}
\thanks{D. Casbeer is with the  Autonomous Control Branch,
Air Force Research Laboratory,
Wright-Patterson A.F.B., OH 45433.
}%
}

\begin{document}
\maketitle

\begin{abstract}
Persistent monitoring missions require an up-to-date knowledge of the changing state of the underlying environment. UAVs can be gainfully employed to continually visit a set of targets representing tasks (and locations) in the environment and collect data therein for long time periods. The enduring nature of these missions requires the UAV to be regularly recharged at a service station. In this paper, we consider the case in which the service station is not co-located with any of the targets. An efficient monitoring requires the revisit time, defined as the maximum of the time elapsed between successive revisits to targets, to be minimized. Here, we consider the problem of determining UAV routes that lead to the minimum revisit time. The problem is NP-hard, and its computational difficulty increases with the fuel capacity of the UAV. We develop an algorithm to construct near-optimal solutions to the problem quickly, when the fuel capacity exceeds a threshold. We also develop lower bounds to the optimal revisit time and use these bounds to demonstrate (through numerical simulations) that the constructed solutions are, on an average, at most 0.01 \% away from the optimum.
\end{abstract}

\section{Introduction} \label{sec:intro}

In this paper, we consider persistent monitoring applications such as forest fire monitoring and military surveillance, in which an environment must be continually monitored to maintain an up-to-date knowledge of its changing state. The environment is represented by a set of distinct target locations of equal priority. The monitoring process involves a UAV visiting these targets to collect data therein, and it is interrupted only by trips to a service station (depot) when the UAV is on the verge of running out of fuel.  A timely acquisition of information can help (for example, in detecting a forest fire or a potential threat to the safety of infrastructure), giving sufficient time to take necessary counter measures. Towards this end, the maximum of the time between successive visits to every target must be minimized; this is achieved by minimizing the maximum taken over all the targets. This maximum is defined as the \emph{revisit time}.

Clearly, the routes taken by the UAV and the location of the service station determine the revisit time. In several applications, the service station is distinct and remote from the target locations. We consider such applications here, and plan routes for the UAV to visit targets and the service station (when necessary) so that the revisit time is minimized.
Arguably, the timing of the visits to the service station is dictated by the fuel capacity of the UAV. As discussed in our prior related work \cite{hari20tro}, a proxy to the fuel capacity is the number of visits, $k$, made by the UAV, at the end of which the UAV must be re-fueled. A systematic procedure to estimate $k$ from the fuel capacity of the UAV is discussed in \cite{hari20tro}. A $k$-visit walk of a UAV is a trip starting from and returning to a depot and includes $k-1$ visits to the set of targets while visiting each target at least once; the $k^{th}$ visit of the UAV is made to the depot. Given that there are $n$ targets, clearly, $k \ge n+1$. To ensure persistent monitoring of targets, a $k-$visit walk is repeated until the end of the mission. Then, the problem of interest, referred to as {\prob} (persistent monitoring problem with a distinct depot), can be stated as follows:

\emph{Given that the UAV is recharged after exactly $k\geq n+1$ visits, find a  $k-$visit walk, which when repeated, has the minimum revisit time}. 
Such a sequence of visits is referred to as an optimal walk and is denoted by $\mathcal{WD}^*(k)$; its revisit time is referred to as the optimal revisit time and is denoted by $\mathcal{RD}^*(k)$.

Given a $k-$visit walk and when $k:=n+1$, the revisit time of the walk is equal to the time required to travel through all the visits in the walk. As a result, the PMP-D is a generalization of the Traveling Salesman Problem (TSP) and is NP-Hard. 
Therefore, the computation time required for finding an optimal solution for PMP-D may not scale well with $n$ and $k$ for every instance of the problem. In this paper, we propose methods to swiftly compute near-optimal solutions to the problem for large values of $k$. We also provide tight bounds to the optimal revisit time, which are helpful in evaluating the quality of heuristic solutions (routes) to this problem. 

 

\subsection{Relevant Literature}
\label{sec:lit-rev}

Route planning in persistent monitoring missions is different from other search or exploration missions due to the requirement of repeatedly visiting certain targets in the environment. With the aim of obtaining an up-to-date knowledge of the changing information, the problem of finding routes with the least revisit time was first proposed in \cite{machado2002multi}. Since then, several researchers have worked on the problem of finding an optimal sequence of visits for persistent monitoring of targets (with equal or unequal priorities) using single or multiple UAVs \cite{chevaleyre2004theoretical,smith,pasqualetti2012complexity,cassandras2011optimal,ccta2018,magesh2011persistent,scherer2016persistent,manyam2017multi,wang2018optimal,las2013persistent,tuyishimire2017cooperative}.  

However, due to the computational complexity of the problem, most of the prior work has ignored the limited fuel/charge capacity of the UAV, the time required to service the UAV, and the location of the service station while planning UAV routes. References \cite{nigam2012control} and \cite{tiwari2018} highlight the importance of accounting for these factors in planning effective persistent monitoring routes. Utilizing a sequence of visits that neglect the servicing requirements of the UAV can result in higher revisit times in practice, as a practitioner may choose to service the UAV at arbitrary timings. This can also result in the UAV spending more time traveling to and from the servicing station as opposed to monitoring the targets. Therefore, to address these issues, it is imperative to focus on planning optimal/near-optimal routes corresponding to different fuel/charge capacities ($k$) of a UAV.

\subsection{Prior work on a special case}
In our prior work \cite{hari20tro}, we considered a special case of {\prob} where the service station is co-located with one of the target locations. We referred to this simpler version\footnote{Every $k-$ visit PMP walk with non-zero service time can be perceived as a feasible $k+2-$visit {\prob} walk (by replacing the service time with visits to and from the depot).} as the persistent monitoring problem (PMP). Just like how small differences in the formulation of the routing problems such as the TSP and the Hamiltonian Path Problem (HPP) ($i.e.$, where the path ends at a node that is distinct from the depot or service station) lead to significant differences in the development of approximation algorithms for the TSP and HPP \cite{svensson2020constant,lam2008traveling}, including a service station that is distinct from any target also leads to several technical challenges in solving the PMP-D. First, PMP-D allows only one visit to the service station in each servicing cycle, therefore, just repeatedly concatenating the Traveling Salesman type tours as done in \cite{hari20tro} will not work for the current problem. Second, as the service station is only visited for recharging, the revisit time of the service station is not considered in the objective of the PMP-D unlike the previous work. These differences significantly change the structure of the optimal solutions for the PMP-D, which we characterize and investigate in this paper. Furthermore, the development of lower bounds on optimal revisit times for persistent monitoring problems requires analyzing revisit sequences (introduced in Section \ref{subsubsec:rs}), $i.e.$, sequences of visits made by the UAV between revisits to a particular target. Due to the presence of a distinct service station, the {\prob} has two different types of revisit sequences, which increases the complexity of analysis.

 Finally, solving {\prob} also helps in tackling an important, but much harder generalization of the the persistent monitoring problem in which targets are prioritized differently. Indeed, the service station in {\prob} can be considered as an additional target with negligible priority (and thereby visited only once). Therefore, by characterizing optimal {\prob} solutions and by introducing weighted targets one at a time, one can provide insights into solving the weighted generalization. 
 
 In \cite{hari20tro}, we characterized the structure of optimal solutions for the special case, and developed methods that exploit this structure thereby leading to a fast computation of near-optimal (and in most instances, optimal) solutions. The computation time{\footnote{The computations were performed on a MacBook Pro with 16 GB RAM and 2.5 GHz Intel Core i7 processor. The optimization solver used was Gurobi.}} was improved from tens of hours to less than a second for instances with no more than 30 targets. Motivated by these results, here, we set out to characterize the structure of optimal solutions for the {\prob}.

\subsection{Contributions \& Organization}

\subsubsection{Contributions}
In this paper, we first develop tight lower bounds to the optimal revisit times corresponding to different values of $k$. Second, utilizing these lower bounds, we develop efficient procedures for constructing optimal/near-optimal feasible solutions to the problem for large values of $k$, for which the computation of optimal solutions is expensive; specifically, we construct feasible solutions for the case $k \geq n^2+2n+1$. Finally, we present the results of extensive numerical simulations for 300 instances of the problem, with $n+1$ ranging from 4 to 50. The results indicate that the revisit times of the constructed feasible solutions are, on average, only 0.01\% away from the lower bounds. In other words, the proposed solutions are either optimal or near-optimal. Furthermore, the time required for computing these solutions is less than a second for instances with no more than $30$ targets and is of the order of a few seconds for larger instances.

\subsubsection{Organization}
In Section \ref{sec:term}, we provide a formal definition of a walk, which includes the restrictions on the permitted sequence of visits. In the same section, we rephrase the definition of revisit time and travel time with regards to the walks. Then, we present a few properties of the revisit times and the travel times of the walks in Section \ref{sec:lemmas}. Next, in order to obtain lower bounds on the optimal revisit times, we divide the set of all walks corresponding to a given $n$ and $k$ into certain categories (the criteria of categorization will be discussed in Section \ref{sec:lb}). Then, utilizing the aforementioned properties of revisit times and travel times, we provide lower bounds on the revisit times of the walks in each category. Finally, we unify the bounds corresponding to each category to obtain a lower bound on the optimal revisit time for a given $k$.

Subsequently, in Section \ref{sec:heuristics}, we utilize these lower bounds to construct optimal/near-optimal feasible walks for large values of $k$. To aid this construction process, for a given $k$, we identify a walk with a smaller number of visits that has a revisit time equal to or almost equal to the lower bound to the optimal revisit time. Then, we populate copies of this smaller walk and perform certain operations on these walks such as addition/removal of visits, altering the starting point of the walk, etc. We introduce these operations in Section \ref{sec:term}. Then, we combine these modified copies to form a walk with $k$ visits such that its revisit time either matches or nearly equals the developed lower bound. If the values match, the obtained walk will be optimal. Otherwise, the gap in the revisit times represents the optimality gap. In Section \ref{sec:lemmas}, we present systematic procedures that aid in combining smaller walks to construct larger walks without increasing the revisit time. Then, in Section \ref{sec:heuristics} we utilize these procedures to present feasible solutions for different values of $k$.


Finally, in Section \ref{sec:num_sim}, we present extensive numerical simulations to demonstrate the quality of the proposed feasible solutions and lower bounds. 

\section{Terminology}
\label{sec:term}




\subsection{Permitted Sequences of Visits}
The servicing needs of the UAV imposes certain restrictions on the sequence in which the UAV can visit the targets and the depot. Throughout the paper, we refer to the set of targets that need to be monitored by $\mathcal{N}$, i.e, $\mathcal{N} =\{1, 2, \cdots, n\}$, and the depot by $d$. We assume that $n \geq 3$ and that the travel time between any pair of targets and between any target and the depot obeys the triangle inequality. Given that the UAV is recharged at the end of every $k$ visits, its permitted sequence of visits in a single servicing cycle is referred to as a \textit{walk}. 

\subsubsection{Walk} A sequence, $(v_0,v_1,v_2, \ldots, v_{k-1}, v_k)$, is defined as a walk with $k$ visits (or simply, a $k-$visit walk) if it satisfies the following conditions: (i) $v_0 = v_k$; (ii) $v_{i-1} \neq v_{i}, \forall i \in \{1,2,\dots,k\}$; and (iii) $\mathcal{N} \subseteq \cup_{i = 1}^{k}  v_i \subseteq \mathcal{N} \cup \{d\} $. The element $v_0$ indicates the starting point of the UAV, and $v_i$ represents the $i^{th}$ visit made by the UAV, $\forall i \in \{1,2, \dots k\}$. To execute a persistent monitoring mission, the walk is repeated for several cycles. The time required by the UAV to travel through all the elements of a walk (in a single servicing cycle) is referred to as the \textit{travel time of the walk}.

A walk with $k$ visits that satisfies the additional conditions $v_0 = v_k=d$ and $\cup_{i = 1}^{k-1}  v_i = \mathcal{N}$ is referred to as a \textit{{\prob} walk}, and is denoted by $\mathcal{WD}(k)$. Similarly, a walk with $k$ visits that satisfies the condition $\cup_{i = 0}^{k}  v_i = \mathcal{N}$ is referred to as a \textit{PMP walk} and is denoted by $\mathcal{W}(k)$.
Note that by definition, $k \geq n+1$ for a {\prob} walk and $k \geq n$ for a PMP walk.

\subsection{Revisit Time}
To adapt the definition of revisit time to a walk, we are interested in sub-sequences of a walk that represent the visits made by the UAV between successive revisits to a target. As walks are repeated for several cycles, we are also interested in sub-sequences that pass through the depot (those that are obtained by repeating the walk). To characterize such sub-sequences, we introduce the term \textit{revisit sequence}.

\subsubsection{Revisit Sequence}
\label{subsubsec:rs}
\begin{figure}
    \centering
    \includegraphics[scale = 0.4]{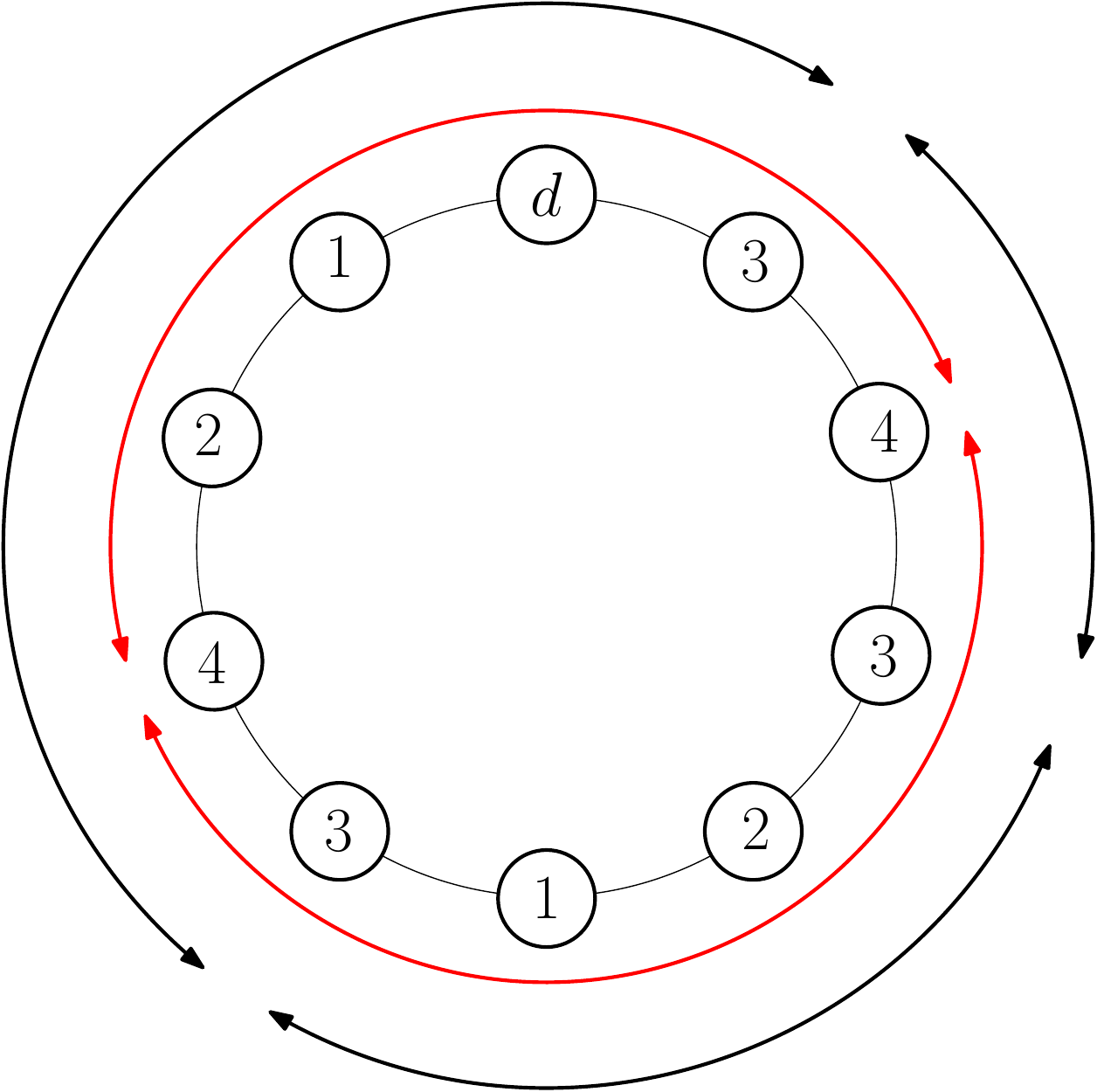}
    \caption{Figure illustrates a {\prob} walk, $\mathcal{WD}(10) = (d, 3, 4, 3, 2, 1, 3, 4, 2, 1, d )$, with $10$ visits to the set of targets, $\mathcal{N} = \{1, 2, 3, 4\}$, and the depot. The revisit sequences of $\mathcal{WD}(10)$ that have target $4$ as the terminus are $(4, 3, 2, 1, 3, 4)$ and $(4, 2, 1, d, 3, 4)$ (indicated in red). The revisit sequences that have target $3$ as the terminus are $(3, 4, 3)$, $(3, 2, 1, 3)$, and $(3, 4, 2, 1, d, 3)$ (indicated in black).}
    \label{fig:revisit_sequences}
\end{figure}

A sequence of $r$ $(\leq k)$ visits of the form $(v_0, v_1, \ldots, v_{r-1}, v_0)$ is referred to as a revisit sequence if it satisfies the following conditions: (i) $v_0 \in \mathcal{N}$ (i.e., $v_0$ is a target and not the depot); (ii) $v_i \in \mathcal{N} \cup \{d\} \setminus \{v_0\}, \forall i \in \{1,2,\dots,r-1\}$; (iii) $v_{i} \neq v_{i+1}, \forall i \in \{1,2,\dots,r-2\}$; and (iv) the depot appears at most once in the sequence. $v_0$ is referred to as the \emph{terminus} of the revisit sequence. The travel time of a revisit sequence represents the time elapsed between successive revisits to the target $v_0$. A revisit sequence that includes visits to all the \textit{targets} is referred to as a \textit{spanning revisit sequence}. Note that a spanning revisit sequence need not include a visit to the depot. Figure \ref{fig:revisit_sequences} depicts the revisit sequences of a walk with targets $3$ and $4$ as the terminus. Among the depicted sequences, $(3, 4, 3)$ and $(3, 2, 1, 3)$ are not spanning, but the rest are spanning.

A walk can be considered as a combination of its revisit sequences. For example, the walk $\mathcal{WD}(10) = (d, 3, 4, 3, 2, 1, 3, 4, 2, 1, d )$ shown in Figure \ref{fig:revisit_sequences} can be considered as a combination of its revisit sequences that have target $3$ as the terminus, i.e., $(3, 4, 3)$, $(3, 2, 1, 3)$, and $(3, 4, 2, 1, d, 3)$; the same applies for revisit sequences with other targets as the terminus.

\subsubsection{Revisit Time of a Walk} The revisit time of a walk is defined as the maximum of the travel times of all its revisit sequences; due to triangle inequality, it is sufficient to consider only spanning revisit sequences.

\subsection{Operations on a Walk}
It will be seen in Section \ref{sec:lemmas} that under certain special conditions, the revisit time of a walk can be related to its travel time. In these cases, we exploit the triangle inequality to develop lower bounds on the revisit/travel time of a walk by skipping one or more visits from the walk. 

\subsubsection{Shortcutting a Walk}
Skipping one or more visits from a given walk to form a \textit{feasible} PMP or {\prob} walk is referred to as shortcutting the given walk; the walk obtained after shortcutting is referred to as a shortcut walk. The requirement that the shortcut walk must be a feasible walk imposes certain restrictions on the visits that can be shortcut (or skipped). For example, a target that is visited exactly once cannot be shortcut. The depot alone cannot be shortcut if the targets visited before and after the depot are the same.




To construct feasible solutions, we often consider walks with smaller number of visits and then combine them to form larger walks.
\subsubsection{Concatenation} Concatenation of two walks $\mathcal{W}_1(k) = (v_0, v_1,v_2, \ldots,v_{k-1}, v_0)$ and $\mathcal{W}_2(m) = (v_0, u_1,u_2, \ldots,u_{m-1}, v_0)$ is defined as $\mathcal{W}_1 \circ \mathcal{W}_2 (k+m)= (v_0, v_1, v_2, \ldots, v_{k-1}, v_0, u_1, u_2, \ldots, u_{m-1}, v_0)$. Note that two walks can be concatenated only when their terminal elements are the same.

To permit concatenation between walks with different terminal elements, we introduce the concept of permutation.



\subsubsection{Permutation}: Permutation of a {\prob} or a PMP walk, $\mathcal{W}(k) = (v_0, v_1, \ldots, v_{r-1},v_r, v_{r+1}, \ldots, v_{k-1}, v_0 )$, about $v_r$ is defined as $\mathcal{C}(\mathcal{W}(k), v_r) = (v_r,v_{r+1}, \ldots, v_{k-1}, v_0, v_1, \ldots, v_{r-1}, v_r)$. Note that the revisit time of a walk is same as that of its permutation, as they have the same revisit sequences.

\subsection{Notation}
To facilitate a compact presentation of the subsequent material, we introduce the following notation.
\begin{itemize}
    \item $\mathcal{R}(\mathcal{W}(k))$ - revisit time of a PMP walk, $\mathcal{W}(k)$.
    \item $\mathcal{R}^*(k)$ - the least revisit time possible for a PMP walk with $k$ visits.
    \item $\mathcal{RD}(\mathcal{WD}(k))$ - revisit time of a {\prob} walk, $\mathcal{WD}(k)$.
    \item $\mathcal{RD}^*(k)$ - the least revisit time possible for a {\prob} walk with $k$ visits.
    \item $\mathcal{W}^*(k)$ - a PMP walk with $k$ visits that has the least revisit time, i.e., $\mathcal{R}(\mathcal{W}^*(k)) = \mathcal{R}^*(k)$. It is referred to as an optimal PMP walk.
    \item $\mathcal{WD}^*(k)$ - a {\prob} walk with $k$ visits that has the least revisit time, i.e., $\mathcal{RD}(\mathcal{WD}^*(k)) = \mathcal{RD}^*(k)$. It is referred to as an optimal {\prob} walk.
    \item $\mathcal{T}(S)$ - travel time of a sequence (any sequence including walks), $S$.
    \item $\mathcal{T}^*(r)$ - minimum travel time possible for a revisit sequence with $r$ visits that does not include the depot.
    \item $\mathcal{TD}^*(r)$ - minimum travel time possible for a revisit sequence with $r$ visits that includes the depot.
\end{itemize}



\section{Properties of $\mathcal{T}^*$, $\mathcal{TD}^*$, $\mathcal{R}$,  and $\mathcal{RD}$}
\label{sec:lemmas}

In this section, we present a few lemmas useful for providing tight bounds on the optimal revisit time; the first three lemmas aid in developing lower bounds, whereas Lemma \ref{im:concat} aids in providing upper bounds.

The first lemma presents the monotonicity property of the minimum travel times of revisit sequences.
\begin{lemma}
$\mathcal{T}^*(v+1) \geq \mathcal{T}^*(v)$, $\forall v \geq n$ and $\mathcal{TD}^*(v+1) \geq \mathcal{TD}^*(v)$, $\forall v \geq n+1$.
\label{im:monotonic_t}
\end{lemma}

\IEEEproof
Let $S$ be any revisit sequence that does not include a visit to the depot; by definition, $v \ge n$ and  $\mathcal{T}(S) \ge \mathcal{T}^*(v)$. Consider an optimal spanning revisit sequence $S^*$ corresponding to $(v+1)$ visits. As $v+1 \geq n+1$, there is at least one target, $t$, that is visited more than once in $S$. Now, consider a sequence $SS$ obtained by shortcutting a visit to $t$ from $S$ to get a revisit sequence of $v$ visits. By triangle inequality, $\mathcal{T}^*(v+1) \ge \mathcal{T}(SS)$; moreover, by definition, $\mathcal{T}(SS) \ge \mathcal{T}^*(v)$. Combining the two inequalities, we get $\mathcal{T}^*(v+1) \geq \mathcal{T}^*(v)$, $\forall v \geq n$.

Consider an optimal spanning revisit sequence that includes a visit to the depot and has $v+1$ visits, where $v \geq n+1$; employing the above shortcutting argument, we get $\mathcal{TD}^*(v+1) \geq \mathcal{TD}^*(v)$, $\forall v \geq n+1$.
\endIEEEproof

The following lemma provides a relation between the least travel times of spanning revisit sequences with and without the depot.
\begin{lemma}
$\mathcal{TD}^*(k+c) \geq \mathcal{T}^*(k)$, $\forall k \geq n$, $c \geq 2$, where $k,c \in \Z_+$; $\mathcal{TD}^*(n+1) \geq \mathcal{T}^*(n)$ .
\label{im:d_to_nd}
\end{lemma}

\IEEEproof
Consider an optimal revisit sequence, $SD$, with $k+c$ visits that includes the depot (i.e., $SD$ has the least travel time among all such revisit sequences). Then, we have $\mathcal{T}(SD) = \mathcal{TD}^*(k+c)$. We prove the lemma by shortcutting the depot from $SD$ and considering the two cases: 1) the targets visited immediately before and after the depot in $SD$ are different; 2) the targets visited immediately before and after the depot in $SD$ are the same.

When the targets visited immediately before and after the depot are the same, shortcutting only the depot leads to an infeasible revisit sequence. However, one can simultaneously shortcut visits to the depot and to the target visited prior to the depot. This results in a feasible revisit sequence, $S_1$, with $k+c-2$ visits with none of the visits being to the depot. Then, because $S_1$ is a shortcut sequence of $SD$, we have $\mathcal{T}(SD) \geq \mathcal{T}(S_1)$. As $S_1$ is a feasible revisit sequence with $k+c-2$ visits and no visits to the depot, we have $\mathcal{T}(S_1) \geq \mathcal{T}^*(k+c-2)$. Additionally, we have from Lemma \ref{im:monotonic_t} that $\mathcal{T}^*(k+c-2) \geq \mathcal{T}^*(k)$, whenever $c \geq 2$. Therefore, combining the inequalities, it follows that $\mathcal{T}^*(k+c) = \mathcal{T}(SD) \geq \mathcal{T}^*(k)$. 

When the targets visited immediately before and after the depot are different, one can shortcut the visit to the depot from $SD$ to form a feasible revisit sequence, $S_2$, that has $k+c-1$ visits, none of which are to the depot. As $S_2$ is a shortcut sequence of $SD$, it follows from the triangle inequality that $\mathcal{T}(SD) \geq \mathcal{T}(S_2)$. Furthermore, as $S$ is a feasible revisit sequence with $k+c-1$ visits and no visits to the depot, we have $\mathcal{T}(S_2) \geq \mathcal{T}^*(k+c-1)$. Additionally, we have from Lemma \ref{im:monotonic_t} that $\mathcal{T}^*(k+c-1) \geq \mathcal{T}^*(k)$, whenever $c \geq 1$. Therefore, combining the inequalities, it follows that $\mathcal{T}^*(k+c) = \mathcal{T}(SD) \geq \mathcal{T}^*(k)$.

When $k = n$ and $c = 1$, every target is visited exactly once in $SD$. Therefore, the argument presented in the previous paragraph holds true, and we have $\mathcal{TD}^*(n+1) \geq \mathcal{T}^*(n)$.
\endIEEEproof

The next lemma relates the revisit times of walks with their travel times, when the number of visits in the walks is at most $2n-1$ or $2n$.

\begin{lemma} Let $\mathcal{W}(v)$ be a PMP walk with $v$ visits, where $v \geq n$, and $\mathcal{WD}(k)$ be a {\prob} walk with $k$ visits, where $k \geq n+1$. Then,
\begin{itemize}
    \item $\forall v \leq 2n-1$, \;  $\mathcal{R}(\mathcal{W}(v)) = \mathcal{T}(\mathcal{W}(v))$, and $\mathcal{R}^*(v) = \mathcal{T}^*(v)$;
    \item $\forall k \leq 2n$, \;  $\mathcal{RD}(\mathcal{WD}(k)) = \mathcal{T}(\mathcal{WD}(k))$, and $\mathcal{RD}^*(k) = \mathcal{TD}^*(k)$. 
\end{itemize}
\label{im:initial_monotonicity}
\end{lemma}

\IEEEproof
Clearly, PMP walks with at most $2n-1$ visits and {\prob} walks with at most $2n$ visits have at least one target that is visited exactly once. Consider {permutations} of these walks about one such target (can be different for different walks). Then, it can clearly be seen that such walks are revisit sequences of themselves. Furthermore, due to the triangle inequality, no other sequences in the walks have a greater travel time than the walk itself. Now, recall that by definition, the revisit time of a walk is equal to the maximum of the travel times of all its revisit sequences. Furthermore, revisit times of walks are not altered by permutations. Therefore, we have that $\mathcal{R}(\mathcal{W}(v)) = \mathcal{T}(\mathcal{W}(v))$, $\forall v \leq 2n-1$, and $\mathcal{RD}(\mathcal{WD}(k)) = \mathcal{T}(\mathcal{WD}(k))$, $\forall k \leq 2n$.

Every spanning revisit sequence is a feasible walk, and a permutation of every feasible PMP walk with at most $2n-1$ visits and every {\prob} walk with at most $2n$ visits is a feasible and spanning revisit sequence. Furthermore, from the previous paragraph, we have that the travel times of these walks are equal to their revisit times. Therefore, it follows that $\mathcal{T}^*(v) = \mathcal{R}^*(v), \forall v \leq 2n-1$, and $\mathcal{RD}^*(k) = \mathcal{TD}^*(k), \forall k \leq 2n$.
\endIEEEproof
The next lemma presents a way to construct larger walks from smaller ones without increasing their revisit times.

\begin{lemma}
\begin{enumerate}
    \item Suppose $k \le 2n-1$. The revisit time of a walk obtained by concatenating a $k-$visit walk with itself or its shortcut walk any number of times is equal to the revisit time of the $k-$visit walk. \label{im:concat-1}
    \item Consider two walks $\mathcal{W}_a$ and $\mathcal{W}_b$ with at most $2n-1$ visits. Let $\mathcal{W}_s$ be a shortcut walk of both $\mathcal{W}_a$ and $\mathcal{W}_b$. Then, the revisit time of a walk obtained by concatenating the three walks any number of times in any sequence such that $\mathcal{W}_a$ and $\mathcal{W}_b$ are not adjacent to each other is equal to the maximum of the revisit times of $\mathcal{W}_a$ and $\mathcal{W}_b$. \label{im:concat-2}
\end{enumerate}
\label{im:concat}
\end{lemma}

\IEEEproof
A detailed proof for Item \ref{im:concat-1} of the lemma can be found in \cite{hari20tro}. 

In Item \ref{im:concat-2}, when $\mathcal{W}_a$, $\mathcal{W}_s$, and $\mathcal{W}_b$ are concatenated such that  $\mathcal{W}_a$ and $\mathcal{W}_b$ are not adjacent to each other, the newly formed revisit sequences belong to the subset of revisit sequences formed by concatenating either $\mathcal{W}_a$ and $\mathcal{W}_s$ or $\mathcal{W}_b$ and $\mathcal{W}_s$. Therefore, following the discussion on Item \ref{im:concat-1}, none of the revisit sequences in the concatenated walk have a travel time larger than the ones in  $\mathcal{W}_a$ and $\mathcal{W}_b$. Henceforth, from the definition of revisit time, it follows that the revisit time of the concatenated walk is the maximum of that of $\mathcal{W}_a$ and $\mathcal{W}_b$.
\endIEEEproof

\section{Lower Bounds on the optimal revisit time}
Given a number $k \geq n+1$, it can always be expressed as $k = pn +q + 1$ for some $p,q \in Z_+$, $p \geq 1$, and $q \leq n-1$.
\label{sec:lb}
\begin{theorem}
\label{thm:lb}
For any $k \ge n+1$,  $\mathcal{RD}^*(k)$ is lower bounded as follows:
\begin{eqnarray*}
\mathcal{RD}^*(k) \ge \begin{cases}
\min\{\mathcal{RD}^*(n+2), \mathcal{R}^*(n+1)\},
\quad \mbox{if} \\ \quad \mathcal{RD}^*(n+1) < \mathcal{R}^*(n+1) \; \;  \& \; \; q = 1;\\
{\mathcal R}^*{(n+1)}, \quad \mbox{if} \\ \quad \mathcal{RD}^*(n+1) < \mathcal{R}^*(n+1) \; \;  \& \; \; q \geq 2; \\
\mathcal{RD}^*(n+1), \quad \mbox{otherwise}.
\end{cases}
\end{eqnarray*}
\end{theorem}

While the lower bounds of Theorem \ref{thm:lb} are valid for any $k \geq n+1$, they are tighter when $k \geq n^2+n+1$. For smaller  $k$, one can construct tighter lower bounds following the solution to the {\color{black}Diophantine Frobenius Problem (DFP) \cite{Frobenius}} and the arguments provided in the proof of Theorem \ref{thm:lb}.


\subsection{Proof for Lower Bound on $\mathcal{RD}^*(k)$ (Theorem \ref{thm:lb})}
\label{sec:lb_proof}

We skip the trivial cases of $k = n+1$ and $k = n+2$, as the bounds clearly hold trivially. For the proof, we may assume without any loss of generality, that $k \ge n+3$. 
We prove the theorem by binning the walks into different categories based on the number of visits in their spanning revisit sequences and showing that the bound holds for each bin. For convenience, we refer to a spanning revisit sequence that has a visit to the depot as r.s.d and a spanning revisit sequence without a visit to the depot as r.s.n.d. Observe that the minimum number of visits in an r.s.d is $n+1$, whereas that in r.s.n.d is $n$. By definition, every {\prob} walk has an r.s.d; while {\prob} may have multiple r.s.ds, note that the depot is only visited exactly once in a {\prob} walk and every r.s.d. 

Given $k \geq n+3$, let us bin the walks based on the number, $v_d$, of visits in their longest (in number of visits) r.s.d. Note that $v_d \geq n+1$. Let $\mathcal{B}(k,n+1)$, $\mathcal{B}(k,n+2)$, and  $\mathcal{B}(k,\geq n+3)$  denote the sets of walks with $v_d = n+1$,  $v_d = n+2$, and $v_d \geq n+3$ respectively. 



\begin{lemma}
Suppose 
$w \in \mathcal{B}(k, \ge n+3)$. Then, $$\mathcal{RD}(w) \ge \max\{\mathcal{RD}^*(n+2), \mathcal{RD}^*(n+1), \mathcal{R}^*(n+1)\}.$$
\end{lemma}
\IEEEproof
Recall that the revisit time of a walk is defined as the maximum of the travel times of all its spanning revisit sequences. Consequently, the travel time of every spanning revisit sequence of a walk serves as a lower bound on the revisit time of the walk. 
Furthermore, every walk in $\mathcal{B}(k,\geq n+3)$ has an r.s.d with at least $n+3$ visits; from Lemma \ref{im:monotonic_t}, it follows that the travel time of such a sequence is at least $\mathcal{TD}^*(n+3)$. Therefore, for every $w \in \mathcal{B}(k,\geq n+3)$, we have $\mathcal{RD}(w) \geq \mathcal{TD}^*(n+3)$.

Next, from Lemma \ref{im:monotonic_t}, we have that $\mathcal{TD}^*(n+3) \geq \mathcal{TD}^*(n+2) \geq \mathcal{TD}^*(n+1)$. Furthermore, from Lemma \ref{im:d_to_nd}, we have that $\mathcal{TD}^*(n+3) \geq \mathcal{T}^*(n+1)$. Finally, from Lemma \ref{im:initial_monotonicity}, we have for $n \geq 3$ that $\mathcal{TD}^*(n+2) = \mathcal{RD}^*(n+2)$, $\mathcal{TD}^*(n+1) = \mathcal{RD}^*(n+1)$, and $\mathcal{T}^*(n+1) = \mathcal{R}^*(n+1)$. Combining the equations and the inequalities in this paragraph with $\mathcal{RD}(w) \geq \mathcal{TD}^*(n+3)$, it follows that  $\mathcal{RD}(w) \geq \mathcal{RD}^*(n+2)$, $\mathcal{RD}(w) \geq \mathcal{RD}^*(n+1)$, and $\mathcal{RD}(w) \geq \mathcal{R}^*(n+1)$. Hence, the theorem is true for all walks with $v_d \geq n+3$.
\endIEEEproof


The rest of the proof is focused on walks in the set $\mathcal{M}(k) := \mathcal{B}(k,n+1) \cup \mathcal{B}(k,n+2)$, i.e., walks with $n+1 \leq v_d \leq n+2$.




\begin{lemma}
\begin{itemize}
\item[(i)] Every target is visited at least twice in any {\prob} walk in $\mathcal{M}(k)$. 
\item[(ii)] Every {\prob} walk in $\mathcal{M}(k)$ has an r.s.n.d.
\end{itemize}
\label{lemma:one-rsnd}
\end{lemma}
\IEEEproof
(i) Consider an arbitrary walk $w \in \mathcal{M}(k)$. Suppose that a target, $v_0 \in \mathcal{N}$, is visited only once in $w$. Then, a permutation of $w$ with $v_0$ as the terminus is the longest r.s.d of $w$, and therefore, we have $v_d = k$. However, this leads to a contradiction as we have $k \geq n+3$ and $v_d \leq n+2$. Hence, every target must be visited at least twice in every walk in ${\mathcal M}(k)$. 

(ii) Refer to the Appendix \ref{app:vd_n2} for a proof.
\endIEEEproof

As a consequence of Lemma \ref{lemma:one-rsnd} (i), we have $k \geq 2n+1$ for all walks in $\mathcal{M}(k)$.

To determine a lower bound, $l(k)$, on the revisit times of all walks in $\mathcal{M}(k)$, we first compute lower bounds, $lb(k,v_d)$, on the revisit times of walks in $\mathcal{B}(k,v_d)$, $v_d \in \{n+1, n+2\}$. Suppose the minimum of revisit times of all walks in $\mathcal{B}(k,v_d)$ is denoted by $\mathcal{RD}^*_B(k,v_d)$. Then, we have $\mathcal{RD}^*_B(k,v_d) \geq lb(k,v_d)$. Thereafter, $l(k)$  can be determined as $l(k) = \min\limits_{n+1 \leq v_d \leq n+2} lb(k,v_d)$.


\subsubsection{\bf Finding $lb(k,vd)$ for the set $\mathcal{B}(k,v_d)$}

Let $w$ be an arbitrary walk in $\mathcal{B}(k,v_d)$. We have from Lemma \ref{lemma:one-rsnd} that $w$ has an r.s.n.d. Let the number of visits in the largest r.s.n.d of $w$ be $v_w$. Let $\min\limits_{w \in \mathcal{B}(k,v_d)}\{v_w\}$ be denoted by $v_{nd}(k,v_d)$. Then, a bound on $\mathcal{RD}^*_{\mathcal{B}}(k,v_d)$ can be obtained by the following lemma.

\begin{lemma}
$\mathcal{RD}^*_{\mathcal{B}}(k,v_d) \geq \max\{\mathcal{TD}^*(v_d), \mathcal{T}^*(v_{nd}(k,v_d))\}$.
\label{lemma:lb(k,vd)}
\end{lemma}

\IEEEproof
By definition, the travel time of every spanning revisit sequence of a walk serves as a lower bound on the revisit time of the walk. Let $w$ be an arbitrary walk in $\mathcal{B}(k,v_d)$. Then, by definition, $w$ has an r.s.d with $v_d$ visits and and r.s.n.d with $v_w$ visits. Further, the travel time of the r.s.d with $v_d$ visits is lower bounded by $\mathcal{TD}^*(v_d)$ and that of the r.s.n.d with $v_w$ visits is lower bounded by $\mathcal{T}^*(v_w)$. Therefore, it follows that $\mathcal{RD}(w) \geq \mathcal{TD}^*(v_d)$ and $\mathcal{RD}(w) \geq \mathcal{T}^*(v_w)$, $\forall{w \in \mathcal{B}(k,v_d)}$. Besides, from Lemma \ref{im:monotonic_t}, it follows that 
$\mathcal{T}^*(v_{nd}(k,v_d))= \min_{w \in \mathcal{B}(k,v_d)}\{\mathcal{T}^*(v_w)\}$. Therefore, we have that $\mathcal{RD}(w) \geq \max\{\mathcal{TD}^*(v_d), \mathcal{T}^*(v_{nd}(k,v_d))\}$, $\forall{w \in \mathcal{B}(k,v_d)}$. Hence, $\mathcal{RD}^*_{\mathcal{B}}(k,v_d) \geq \max\{\mathcal{TD}^*(v_d), \mathcal{T}^*(v_{nd}(k,v_d))\}$.
\endIEEEproof

The previous lemma provides a lower bound on $\mathcal{RD}^*_{\mathcal{B}}(k,v_d)$; however, this bound is dependent on $v_{nd}(k,v_d)$. To obtain an expression for $lb(k,v_d)$ that is independent of $v_{nd}(k,v_d)$, we utilize the following lemma.

\begin{lemma}
$v_{nd}(k,v_d) \geq n+1$ if $n+1 \leq v_d \leq n+q$, and $v_{nd}(k,v_d) = n$ if $v_d = n+q+1$.
\label{lemma:vnd}
\end{lemma}

\IEEEproof
We prove the lemma by showing that in the range $n+1 \leq v_d \leq n+q+1$, $v_{nd}(k,v_d)$ can take the value $n$ only when $v_d = n+q+1$.

Let the set of targets be $\mathcal{N}$ = $\{t_1, t_2, \ldots, t_{n-1},t_n\}$. Let us consider an arbitrary walk $w \in \mathcal{B}(k,v_d)$. From Lemma \ref{lemma:one-rsnd}, we have that $w$ has a spanning r.s.n.d. Therefore, $v_{nd}(k,v_d) \geq n$. Now, let us suppose that  $v_{nd}(k,v_d) = n$. This implies that every spanning r.s.n.d in $w$ has exactly $n$ visits; WLOG, we can assume that one such sequence is $\mathcal{S} = (t_1,t_2, \ldots, t_{n-1},t_n,t_1)$. Then, the next visit in $w$ that follows this sequence must be $t_2$; if not, we get a spanning r.s.n.d with $t_2$ as the terminus (starting from the visit to $t_2$ in $\mathcal{S}$ and ending at the visit to $t_2$ that is outside $\mathcal{S}$) that has at least $n+1$ visits. Similarly, the visit after that must be $t_3$, and so on. Therefore,  $v_{nd}(k,v_d) =n$ implies that $\mathcal{S}$ must get repeated. Note that we can have at most $p$ \textit{full} occurrences of $\mathcal{S}$ because there are a total of $pn+q+1$ visits and every occurrence of $\mathcal{S}$ has exactly $n$ visits. As a consequence, the $p^{th}$ occurrence of $\mathcal{S}$ is followed by only the first $q$ visits from $\mathcal{S}$ and a visit to the depot (note that the depot must be visited exactly once in a {\prob} walk). Then, WLOG, $w = (d,\underbrace{t_1,t_2,\ldots,t_n,t_1}_{\mathcal{S}},\ldots,t_n,t_1, \ldots, \underbrace{t_1,\ldots,t_n,t_1}_{p^{th} \text{ occurrence of }\mathcal{S}},\ldots,t_q,d)$.
Then, the sequence with $t_n$ as the terminus (sequence starting with the last visit to $t_n$ in $w$ and terminating with the first visit to $t_n$ in $w$) has a visit to the depot, spans all the targets and has the highest $v_d$. Note that $v_d = n+q+1$, that is $q$ visits to other targets, followed by a visit to the depot, which is further followed by $n$ visits to targets. Therefore, $v_{nd}(k,v_d)$ takes the value $n$ only when $v_d = n+q+1$. Otherwise, $v_{nd}(k,v_d) \geq n+1$.
\endIEEEproof

Now, following Lemmas \ref{lemma:vnd} and \ref{im:monotonic_t}, we obtain that $\mathcal{T}^*(v_{nd}(k,v_d)) \geq \mathcal{T}^*(n+1)$ when $n+1 \leq v_d \leq n+q$, and $\mathcal{T}^*(v_{nd}(k,v_d)) = \mathcal{T}^*(n)$ when $v_d = n+q+1$. Next, applying these bounds to Lemma \ref{lemma:lb(k,vd)}, we have $\mathcal{RD}^*_{\mathcal{B}}(k,v_d) \geq \max\{\mathcal{TD}^*(v_d), \mathcal{T}^*(n+1)\}$ if $n+1 \leq v_d \leq n+q$, and $\mathcal{RD}^*_{\mathcal{B}}(k,v_d) \geq \max\{\mathcal{TD}^*(v_d), \mathcal{T}^*(n)\}$ if $v_d = n+q+1$.

\textbf{Remark}: 
{Following the aforementioned results, we set the lower bounds on $\mathcal{RD}^*_{\mathcal{B}}(k,v_d)$ to be the following: 1) $lb(k,v_d) = \max\{\mathcal{TD}^*(v_d), \mathcal{T}^*(n+1)\}$ if $n+1 \leq v_d \leq n+q$; and 2) $lb(k,n+q+1) = \max\{\mathcal{TD}^*(v_d), \mathcal{T}^*(n)\}$.

When $k <n^2+n+1$, the bounds in Lemma \ref{lemma:vnd} (for the case $n+1 \leq v_d \leq n+q$) can be further tightened to obtain improved bounds on $\mathcal{RD}^*_{\mathcal{B}}(k,vd)$. However, this is out of the scope of this paper.}

\vspace*{0.1in}

\subsubsection{{\bf Determination of $l(k)$}}

Next, we show that it is sufficient to compute $lb(k,v_d)$ for the range $n+1 \leq v_d \leq \min\{n+2, n+q+1\}$ to determine $l(k)$. 
\begin{lemma}
$lb(k,v_d) \geq lb(k,n+q+1)$, $\forall{v_d \geq n+q+2}$. Therefore, $l(k)$ = $\min \{lb(k,v_d): n+1 \leq v_d \leq \min\{n+2,n+q+1\}\}$. 
\end{lemma}

\IEEEproof
When $v_d = n+q+1$, we have from Lemma \ref{lemma:vnd} that $v_{nd}(k,v_d) = n$. Therefore, $lb(k,n+q+1)$ = $\max\{\mathcal{TD}^*(n+q+1)$, $\mathcal{T}^*(n)\}$. For $v_d \geq n+q+2$, it follows from Lemma \ref{im:monotonic_t} that $\mathcal{TD}^*(v_d) \geq \mathcal{TD}^*(n+q+1)$. Furthermore, by definition, $v_{nd}(k,v_d) \geq n$. Therefore, from Lemma \ref{im:monotonic_t}, we have that $\mathcal{T}^*(v_{nd}(k,v_d)) \geq \mathcal{T}^*(n)$. As a result, we have $lb(k,v_d)$ = $\max\{\mathcal{TD}^*(v_d)$, $\mathcal{T}^*(v_{nd}(k,v_d))\}$ $\geq$ $\max\{\mathcal{TD}^*(n+q+1)$, $\mathcal{T}^*(n)\}$ = $lb(k,n+q+1)$. Hence,  $l(k)$ = $\min\{lb(k,v_d): n+1 \leq v_d \leq n+2\}$ = $\min \{lb(k,v_d):n+1 \leq v_d \leq \min\{n+2, n+q+1\}\}$.

\endIEEEproof

When $q \leq 2$, the values of $lb(k,v_d)$ for the range $n+1 \leq v_d \leq \min\{n+2,n+q+1\}$ are presented in Table \ref{tab:lower_bound_analysis}.

\begin{table}[htp!]
    \centering
    \caption{Lower bounds on the revisit times of walks in the set $\mathcal{B}(k,v_d)$, where $k$ is expressed as $pn+q+1$.}
    \begin{tabular}{|c|c|c|c|c|}
    \toprule
    $q$ & $v_d$ & $v_{nd}(k,v_d)$ &$lb(k,v_d)$\\
    \toprule
    $q = 0$ & $  n+1$ & $n$ & $\max\{\mathcal{TD}^*(n+1)$, $\mathcal{T}^*(n)$\}\\
    \midrule
    \multirow{2}{*}{$q = 1$} & $n+1$ & $ \geq n+1$ & $\max \{\mathcal{TD}^*(n+1), \mathcal{T}^*(n+1) \}$\\
    & $ n+2$ & $n$ & $\max \{\mathcal{TD}^*(n+2), \mathcal{T}^*(n) \}$\\
    \midrule
    \multirow{2}{*}{$q = 2$} & $n+1$ & $\geq n+1$ & $\max \{\mathcal{TD}^*(n+1), \mathcal{T}^*(n+1) \}$\\
     & $n+2$ & $\geq n+1$ & $\max \{\mathcal{TD}^*(n+2), \mathcal{T}^*(n+1) \}$\\
    \bottomrule
    \end{tabular}
    \label{tab:lower_bound_analysis}

\end{table}

Next, observe that the bounds for the case $q>2$ are the same as that for $q = 2$. That is, $\forall{q \geq 2}$, we have  $lb(k, n+1) = \max\{\mathcal{TD}^*(n+1),\mathcal{T}^*(n+1)\}$ and  $lb(k,n+2) = \max\{\mathcal{TD}^*(n+2),\mathcal{T}^*(n+1)\}$. Therefore, the values of $l(k)$ for $q>2$ are the same as that for $q = 2$.

By applying Lemmas \ref{im:monotonic_t}, \ref{im:d_to_nd}, and $l(k) = \min\{lb(k,v_d): n+1 \leq v_d \leq \min\{n+2,n+q+1\}\}$, the values of $l(k)$ reduce to the following:\\
$l(k) = 
\begin{cases}
\mathcal{TD}^*(n+1), \text{ if } q = 0;\\
\min 
     \begin{cases}
     {\max \{\mathcal{TD}^*(n+1), \mathcal{T}^*(n+1) \},  }\\
     \mathcal{TD}^*(n+2)
     \end{cases}, \text{ if } q = 1;\\
\max \{\mathcal{TD}^*(n+1), \mathcal{T}^*(n+1) \}, \text{ if } q \geq 2.
\end{cases}$

As $n \geq 2$, we have from Lemma \ref{im:initial_monotonicity} that $\mathcal{RD}^*(n+2) = \mathcal{TD}^*(n+2)$, $\mathcal{RD}^*(n+1) = \mathcal{TD}^*(n+1)$, and $\mathcal{R}^*(n+1) = \mathcal{T}^*(n+1)$. Substituting these values in $l(k)$ proves the theorem.

\section{Optimal/Near-optimal solutions}
\label{sec:heuristics}
In this section, guided by the lower bounds provided in Section \ref{sec:lb}, we construct feasible {\prob} solutions with $k \geq n^2+2n+1$ visits\footnote{Though we provide upper bounds only for the case $k \geq n^2+2n+1$, simulations suggest that these bounds are also valid for $k \geq n^2+n+1$.} such that their revisit times nearly match or are equal to the lower bounds. It will be shown in Section \ref{sec:num_sim} that these solutions are indeed optimal for all the instances to which optimal solutions were available. Besides, the average percentage gap between upper and lower bounds on the revisit times computed over 300 instances (with $n+1$ ranging from 4 to 50) was 0.01\%.


Given $k$, the idea behind the construction is to start with a smaller walk (at most $n+2$ visits) that has a revisit time equal/nearly-equal to the lower bounds provided in Section \ref{sec:lb}. Utilizing this walk, we derive a set of intermediate walks such that all except one visit in these walks are in the same sequence. Then, using Lemma \ref{im:concat} and the solution to {\color{black}DFP}, we concatenate these walks such that the resultant walk has $k$ visits and its revisit time is the maximum of that of all its concatenated walks.

Note that the lower bounds in Section \ref{sec:lb} are the revisit times of either an optimal PMP walk (over $n$ targets) with $n+1$ visits or an optimal {\prob} walk (over $n$ targets and a depot) with $n+1$ or $n+2$ visits. Therefore, we start with one of these walks, and construct intermediate walks such that the set of walks consist of the following: one {\prob} walk with either $n+1$ or $n+2$ visits; one PMP walk with $n$ visits; and in most cases, one PMP walk with $n+1$ visits. These walks must be constructed such that the PMP walk with $n$ visits is a shortcut walk of the other walks in the set. Since the number of visits in these walks is at most $n+2$ (where $n \geq 3$), there is at least one target, say target $t$, that is visited exactly once in all these walks. Then, consider a permutation of the above walks about $t$, and let them be $\mathcal{WD}^i(n+1 \text{ or }n+2)$, $\mathcal{W}^i(n)$ and $\mathcal{W}^i(n+1)$ in the respective order. Then, concatenate these walks as shown in the following equation to obtain a walk with $k$ visits.

\begin{multline}
    \mathcal{CWD^i}(k) = \mathcal{WD}^i(n+1 \text{ or }n+2) \circ \mathcal{W}^i(n) \circ \\
    \mathcal{W}^i(n \text{ or } n+1) \circ \cdots  \circ \mathcal{W}^i(n \text{ or } n+1)\circ \mathcal{W}^i(n)
    \label{eq:concat-feasible-walk-1}
\end{multline}

Note that for $k \geq n^2+2n+1$, such a construction is always possible following the solution of the {\color{black} DFP}
\footnote{Every integer that is at least $n^2-n$ can be expressed as a non-negative integer conic combination of $n$ and $n+1$.}.
A permutation of $\mathcal{CWD}(k)$ about the depot provides the desired walk, $\mathcal{WD}_{desired}(k)$. That is,
\begin{equation}
    \mathcal{WD}_{desired}(k) = \mathcal{C}(\mathcal{CWD}(k),d).
    \label{eq:concat-feasible-walk-2}
\end{equation}
In the following subsection, we list the sets of intermediate walks useful in the construction process.

\subsection{Sets of intermediate walks}
We denote the sets of walks that lead to the construction of \textit{optimal} solutions as $Oi$, where $i \in \{1,2\}$. The remaining sets are denoted by $Hj$, $\forall j \in \{1, 2, 3\}$, where $H$ stands for \textit{heuristics}. Nonetheless, in certain cases, sets $Hj$, $\forall j \in \{1,2,3\}$, can also lead to the construction of optimal solutions as will be discussed later.
\subsubsection{O1 - $\mathcal{WD}^*(n+1)$ and its {shortcut} walk}
\begin{itemize}
\setlength{\itemindent}{1.5em}
    \item $\mathcal{WD}^*(n+1)$: an optimal {\prob} walk with
    $n+1$ visits
    \item $\mathcal{WD}^1_{SD}$: shortcut the depot from $\mathcal{WD}^*(n+1)$
\end{itemize}

\subsubsection{O2 - $\mathcal{WD}^*(n+2)$ and its {shortcut} walk}
\begin{itemize}
\setlength{\itemindent}{1.5em}
    \item $\mathcal{WD}^*(n+2)$: an optimal {\prob} walk with $n+2$ visits
    \item $\mathcal{WD}^2_{STD}$: shortcut the repeated target and the depot from $\mathcal{WD}^*(n+2)$
\end{itemize}

\subsubsection{H1 -  {shortcut} walks of $\mathcal{WD}^*(n+2)$}
\begin{itemize}
\setlength{\itemindent}{1.5em}
    \item $\mathcal{WD}^2_{ST}$: shortcut the repeated target from $\mathcal{WD}^*(n+2)$
    \item $\mathcal{WD}^2_{SD}$: shortcut the depot from $\mathcal{WD}^*(n+2)$
    \item $\mathcal{WD}^2_{STD}$: shortcut the repeated target and the depot from $\mathcal{WD}^*(n+2)$
\end{itemize}

\subsubsection{H2 - $\mathcal{W}^*(n+1)$ and its derived walks}
Here, in addition to $\mathcal{W}^*(n+1)$ and its shortcut walk, we also consider a walk obtained by adding a visit to the shortcut walk of $\mathcal{W}^*(n+1)$.
\begin{itemize}
\setlength{\itemindent}{1.5em}
    \item $\mathcal{W}^*(n+1)$: an optimal PMP walk with $n+1$ visits
    \item $\mathcal{W}^1_{ST}$: shortcut the repeated target from  $\mathcal{W}^*(n+1)$
    \item $\mathcal{W}^1_{STID}$:
    the derivation of this walk involves forming $n+1$-visit walks by inserting the depot between every two visits of $\mathcal{W}^1_{ST}$. Among the resulting walks, the one with the least travel time is the desired walk.
\end{itemize}

\subsubsection{H3 - $\mathcal{WD}^*(n+1)$ and its {derived} walks}
Here, in addition to $\mathcal{WD}^*(n+1)$ and its shortcut walk, we also consider a walk obtained by adding a visit to the shortcut walk of $\mathcal{WD}^*(n+1)$.
\begin{itemize}
\setlength{\itemindent}{1.5em}
    \item $\mathcal{WD}^*(n+1)$: an optimal {\prob} walk with $n+1$ visits
    \item $\mathcal{WD}^1_{SD}$: shortcut the depot from $\mathcal{WD}^*(n+1)$
    \item $\mathcal{WD}^1_{SDIT}$:
    the derivation of this walk involves forming \textit{feasible} $n+1$-visit walks by inserting a target between every two visits of $\mathcal{WD}^1_{SD}$ and repeating the same with every target. Among the resulting walks, the one with the least travel time is the desired walk.
\end{itemize}

We refer to a solution constructed from the intermediate walks of a set $Oi$ or $Hj$ as a \emph{solution of type $Oi$ or $Hj$} respectively, where $i \in \{1,2\}$, and $j \in \{1, 2, 3\}$.

\subsection{Proposed Solutions}
Similar to the lower bounds, the solutions we propose depend on the value of $k$ (or $q$) and the relative values of $\mathcal{R}^*(n+1)$, $\mathcal{RD}^*(n+1)$, and $\mathcal{RD}^*(n+2)$.

\subsubsection{{$k = pn +1$, i.e., $q=0$}}
When $q = 0$, the solution we propose is of the type $\bm{O1}$, irrespective of the relative values of $\mathcal{RD}^*(.)$ and $\mathcal{R}^*(.)$.

\begin{theorem}
Given a number of visits, $k$, that can be expressed as $pn+1$, where $p \in Z_{++}$, a solution of the type $O1$ with $k$ visits is optimal for {\prob}.
\label{thm:O1-optimal-q0}
\end{theorem}

\IEEEproof
When $k = pn+1$, a feasible {\prob} solution of the type $O1$ can always be constructed by concatenating $p-1$ copies of $\mathcal{WD}^1_{SD}$ with a copy of $\mathcal{WD}^*(n+1)$ as shown in Equations \eqref{eq:concat-feasible-walk-1} and \eqref{eq:concat-feasible-walk-2}. As $\mathcal{WD}^1_{SD}$ is a shortcut walk of $\mathcal{WD}^*(n+1)$, it follows from Lemma \ref{im:concat} that the revisit time of the concatenated walk is equal to the revisit time of $\mathcal{WD}^*(n+1)$, which is equal to $\mathcal{RD}^*(n+1)$. From Theorem \ref{thm:lb}, we have that the lower bound on the optimal revisit time is $\mathcal{RD}^*(n+1)$. Therefore, a solution of the type $O1$ is optimal.
\endIEEEproof


\subsubsection{{$\mathcal{R}^*(n+1) > \mathcal{RD}^*(n+2)$}}
We divide this case into two sub-cases, $q=1$ and $q \geq 2$, for which the lower bounds on $\mathcal{RD}^*(k)$ are $\mathcal{RD}^*(n+2)$ and $\mathcal{R}^*(n+1)$ respectively.\\
\indent \indent {$\mathit{q=1}$ ($\mathit{k = pn+2}$)}: When $q = 1$, the solution we propose is of the type $O2$.

\begin{theorem}
Given a number of visits, $k$, that can be expressed as $pn+2$, where $p \in Z_{++}$, if $\mathcal{R}^*(n+1) > \mathcal{RD}^*(n+2)$, a solution of the type $O2$ with $k$ visits is optimal for {\prob}.
\label{thm:O2-optimal-q1}
\end{theorem}

\IEEEproof
When $k = pn+2$, a feasible {\prob} solution of the type $O2$ can always be constructed by concatenating $p-1$ copies of $\mathcal{WD}^2_{STD}$ with a copy of $\mathcal{WD}^*(n+2)$ as shown in Equations \eqref{eq:concat-feasible-walk-1} and \eqref{eq:concat-feasible-walk-2}. As the $\mathcal{WD}^2_{STD}$ is a shortcut walk of $\mathcal{WD}^*(n+2)$, it follows from Lemma \ref{im:concat} that the revisit time of the obtained walk is equal to $\mathcal{RD}^*(n+2)$. From Theorem \ref{thm:lb}, we have that the lower bound on $\mathcal{RD}^*(k)$ for this case is $\mathcal{RD}^*(n+2)$, which is equal to the revisit time of the constructed solution. Therefore, a solution of the type $O2$ with $k$ visits is optimal.
\endIEEEproof


\indent \indent {$\mathit{q \geq 2}$}: In this case, the lower bound on $\mathcal{RD}^*(k)$ is $\mathcal{R}^*(n+1)$. Thereby, the solution we propose is of the type $H2$, which consists of $\mathcal{W}^*(n+1)$ and its derived walks. The desired walk consists of one copy of $\mathcal{W}^1_{STID}$ concatenated with the required number of copies of $\mathcal{W}^*(n+1)$ and $\mathcal{W}^1_{ST}$. Clearly, the cost of $\mathcal{W}^1_{ST}$ is no more than that of $\mathcal{W}^*(n+1)$. Based on extensive numerical simulations, we propose the following conjecture on the cost of $\mathcal{W}^1_{STID}$.

\begin{conjecture}
\label{conj:swad_cost}
If $\mathcal{R}^*(n+1) > \mathcal{RD}^*(n+2)$, then $T(\mathcal{W}^1_{STID}) \leq T(\mathcal{W}^*(n+1))$.
\end{conjecture}

If Conjecture \ref{conj:swad_cost} is true, the cost of $\mathcal{W}^1_{STID}$ is no more than that of $\mathcal{W}^*(n+1)$ in the current case. Then, from Lemma \ref{im:concat}, it follows that the revisit time of the constructed walk is $\mathcal{R}^*(n+1)$, which is equal to the lower bound on $\mathcal{RD}^*(k)$. Therefore, the \textit{proposed solution is optimal if Conjecture \ref{conj:swad_cost} is true}.


\subsubsection{{$\mathcal{R}^*(n+1) \leq \mathcal{RD}^*(n+2)$}}
As we discussed the case $q=0$ above, here we only discuss the case $q \geq 1$. We further split this case into two subcases based on whether
it is possible to shortcut the depot from $\mathcal{WD}^*(n+2)$.

\indent \textit{$\bullet$ It is possible to shortcut the depot from} $\mathit{\mathcal{WD}^*(n+2)}$: In this case, the proposed solution is the one that has the least revisit times among the solutions of type $H1$, $H2$, and $H3$. Towards this end, we first compute $r_1 = \max\{T(\mathcal{WD}^2_{ST}), T(\mathcal{WD}^2_{SD})\}$, $r_2 = \max\{\mathcal{R}^*(n+1), T(\mathcal{W}^1_{ST})\}$, $r_3 = \max\{\mathcal{RD}^*(n+1), T(\mathcal{WD}^1_{SDIT})\}$. Then, if $\min\{r_1,r_2,r_3\}$ is equal to $r_1$, the proposed solutions is of the type $H1$, if it is $r_2$, the proposed solution is of the type $H2$, and if it is $r_3$, the proposed solution is of the type $H3$. Ties are broken arbitrarily, and the revisit time of the obtained walk is $\min\{r_1, r_2, r_3\}$.

\indent \textit{$\bullet$ It is not possible to shortcut the depot from} $\mathit{\mathcal{WD}^*(n+2)}$: In this case, one cannot construct a solution of the type $H1$ as $\mathcal{WD}^2_{SD}$ does not exist. Therefore, the proposed solution is of the type $H2$ if $r_2 \leq r_3$, and it is of the type $H3$ otherwise, where $r_2$ and $r_3$ are computed as shown in above paragraph.



\begin{figure}
        \includegraphics[scale=0.8]{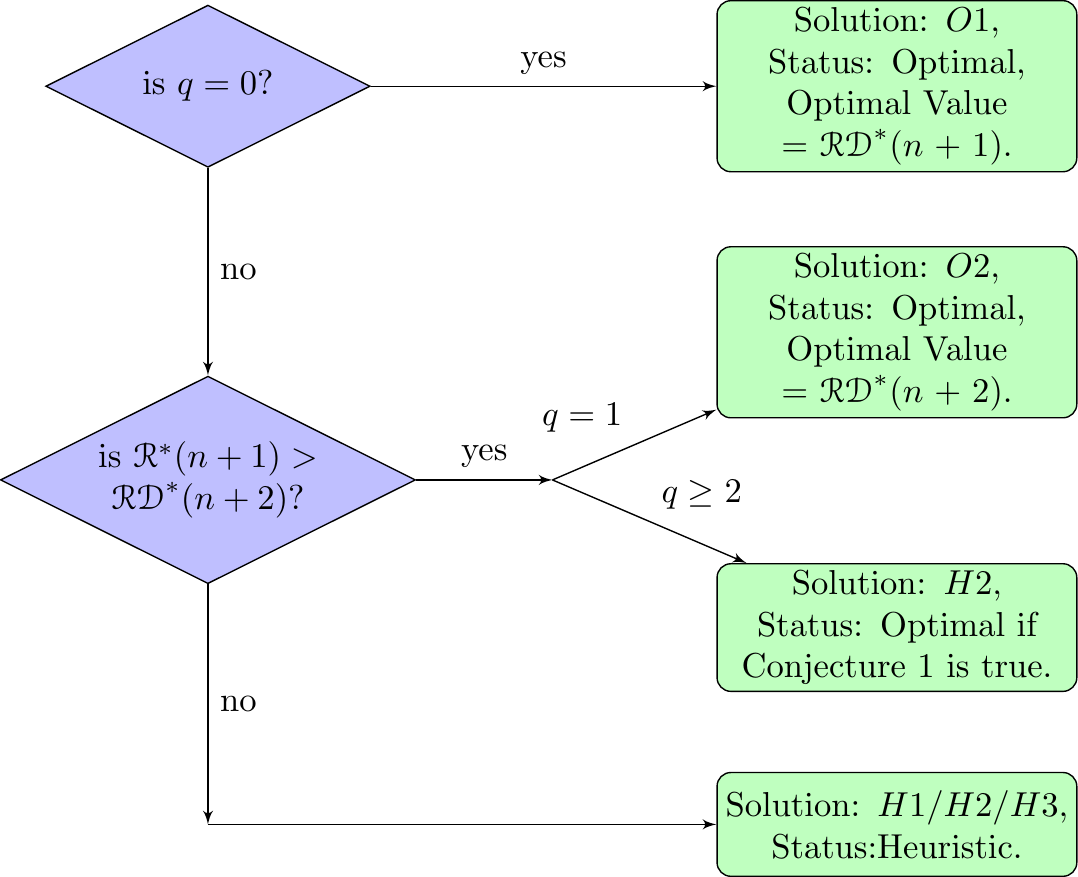}
        \caption{Flowchart summarizing the proposed solutions for different cases}
        \label{fig:flowchart_sol}
    \end{figure}

The solutions proposed for different cases are summarized in the form of a flowchart in Figure \ref{fig:flowchart_sol}

\tikzstyle{decision} = [diamond, aspect = 2, draw, fill=blue!25,
     text width=7em, text badly centered, node distance=4.5cm, inner sep=1pt]
\tikzstyle{block} = [rectangle, draw, fill=green!25, node distance = 1cm,
    text width=10em, text centered, rounded corners, minimum height=3em, inner sep = 2pt]
\tikzstyle{line} = [draw, -latex']

\section{Numerical Simulations}
\label{sec:num_sim}

In this section, we illustrate the structure of optimal {\prob} solutions, evaluate the quality of the upper and lower bounds on the optimal revisit times, and show the effectiveness of the proposed feasible solutions through the results of numerical simulations performed over 300 instances; the number of targets and depot in the instances range from 4 to 50. All the simulations were performed on a Macbook Pro with 16 GB RAM and Intel Core i7 processor with a processing speed of 2.5 GHz.

Given $n$ and $k$, an optimal solution to the {\prob} can be obtained by solving a Mixed Binary Linear Program (MBLP) formulation similar to the one presented in \cite{hari2018persistent}. In this work, we implemented the formulation using Julia \cite{bezanson2017julia} with the help of a package named JuMP \cite{DunningHuchetteLubin2017}. The formulation was solved using the CPLEX solver \cite{cplex2009v12}. It is to be noted that when $k \leq 2n-1$, the revisit time of a walk is equal to its travel time. Therefore one can use standard TSP-type formulations, similar to the Integer Linear Program (ILP) formulation presented in \cite{hari2019icuasform}, to find an optimal walk with at most $2n-1$ visits.

\subsection{Optimal Solutions}
\begin{figure}
    \centering
    \includegraphics[scale=0.75]{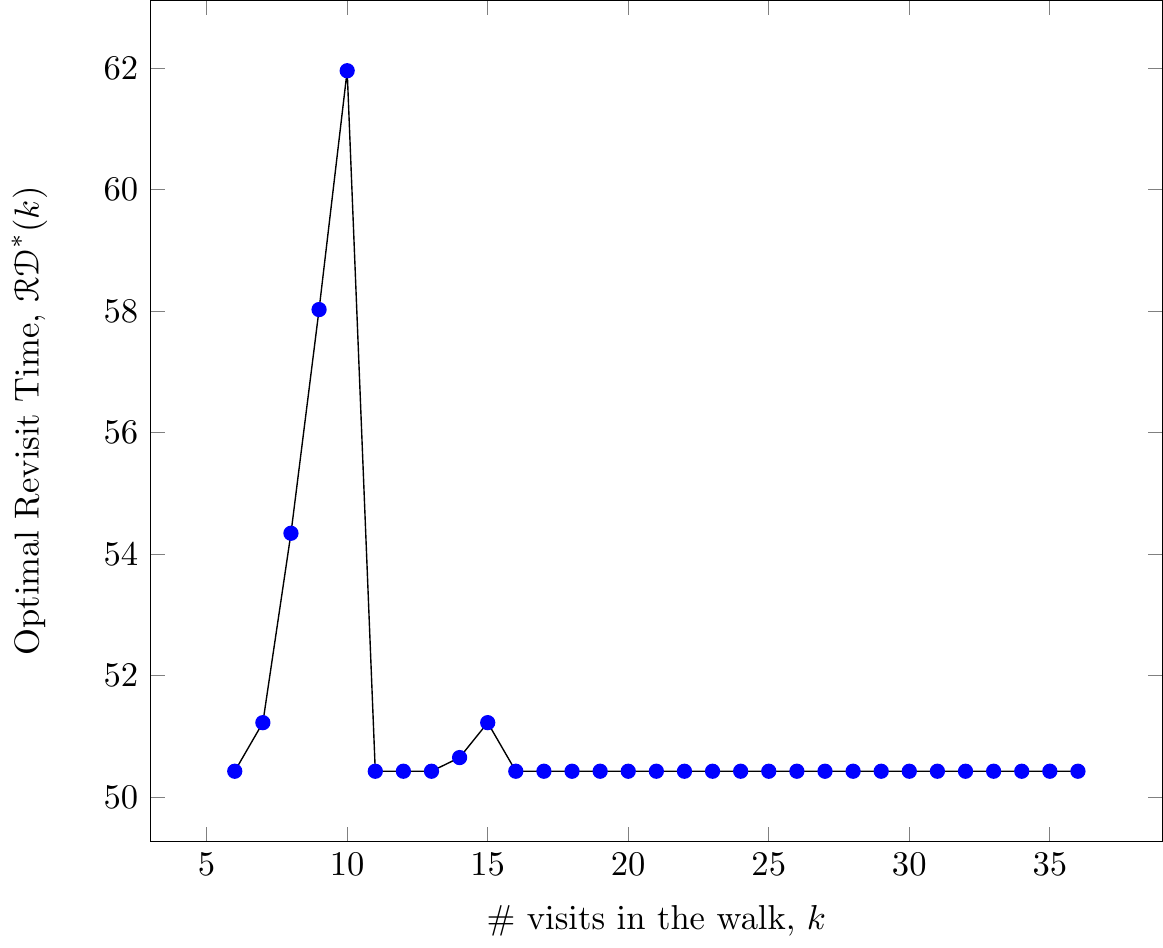}
       \caption{Figure illustrating the plot of the optimal revisit time against the number of visits in the walk for an instance with $5$ targets and a depot. For this instance, optimal revisit time takes exactly one value asymptotically}
    \label{fig:6-4-ort-d}
\end{figure}

\begin{figure}[h!]
   \includegraphics[scale=0.75]{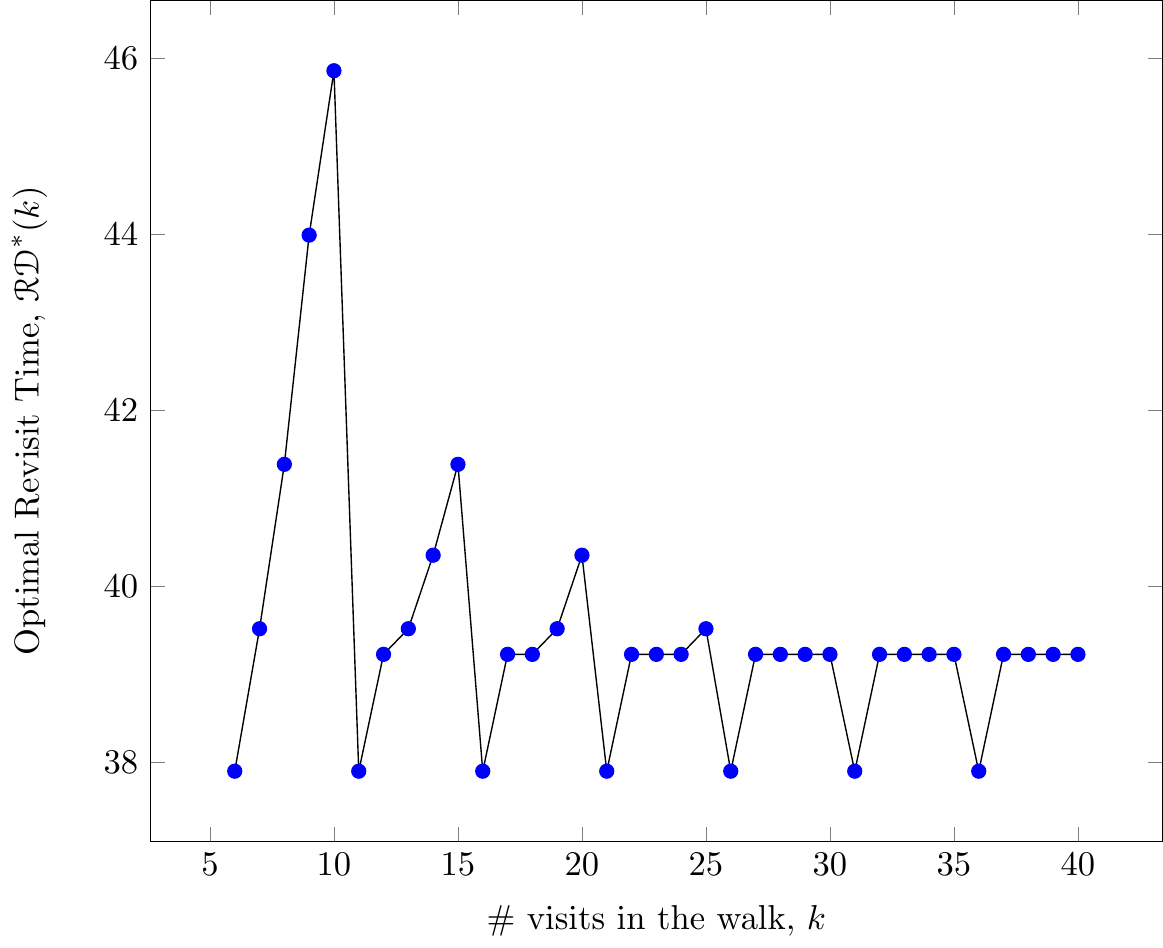}
    \caption{Figure illustrating the plot of the optimal revisit time against the number of visits in the walk for an instance with $5$ targets and a depot. For this instance, optimal revisit time takes two different values asymptotically}
    \label{fig:6-1-ort-d}
\end{figure}

\begin{figure}[h!]
   \includegraphics[scale=0.75]{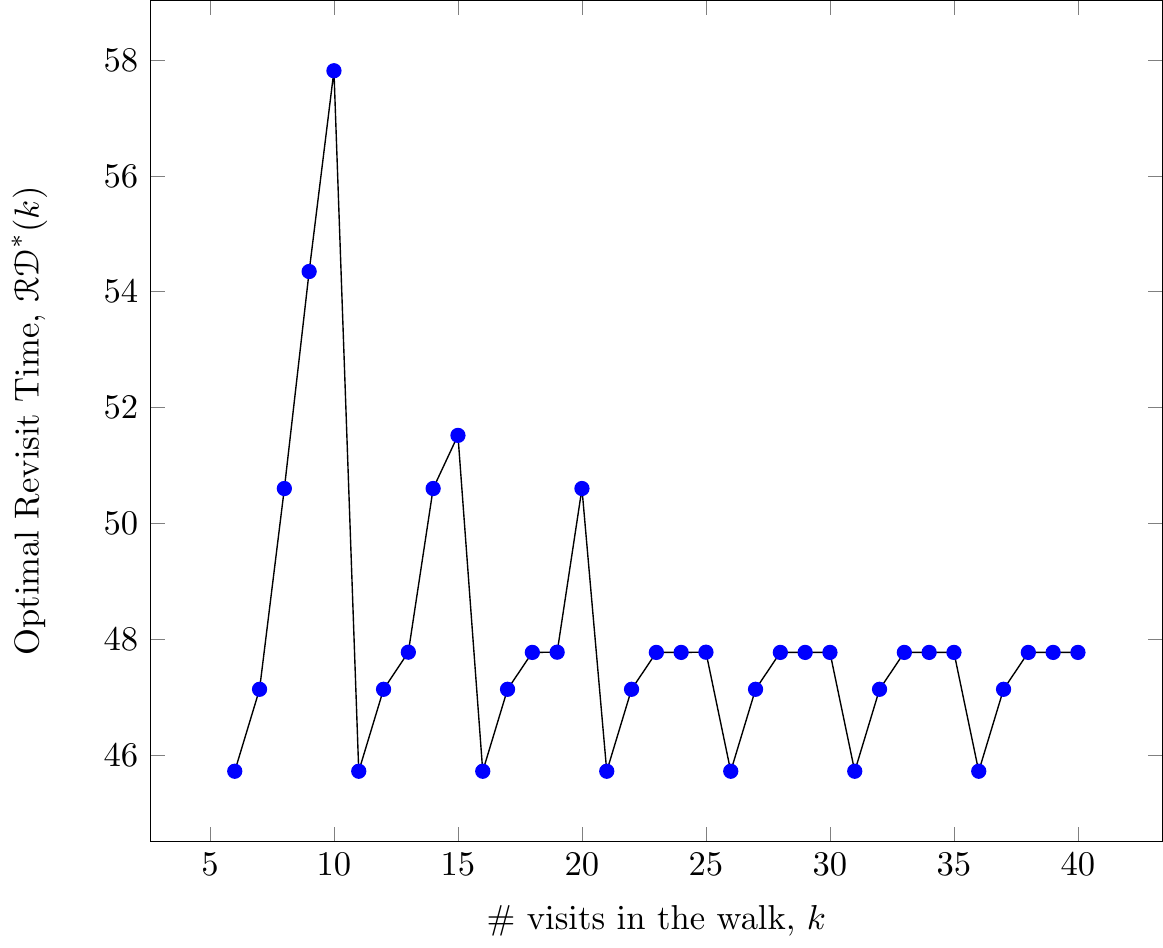}
    \caption{Figure illustrating the plot of the optimal revisit time against the number of visits in the walk for an instance with $5$ targets and a depot. For this instance, optimal revisit time takes three values asymptotically}
    \label{fig:6-21-ort-d}
\end{figure}

Firstly, to examine the optimal solutions, for each instance, we solved the MBLP formulation starting from $k = n+1$ to a value of $k$ for which optimal solutions were computable (on the specified laptop). The computation time was observed to increase rapidly with the values of $n$ and $k$, and optimal solutions with large number of visits could not be computed for instances with $n+1$ beyond 8. For each instance, we computed the optimal revisit times and plotted them against $k$. Sample plots for instances with $n+1 = 6$ are shown in Figures \ref{fig:6-4-ort-d}, \ref{fig:6-1-ort-d}, and \ref{fig:6-21-ort-d}. To aid the discussion presented in earlier sections, for each instance, we also computed an optimal PMP walk with $n+1$ visits and its revisit time, using the ILP formulation.

\begin{figure}
    \centering
    \includegraphics{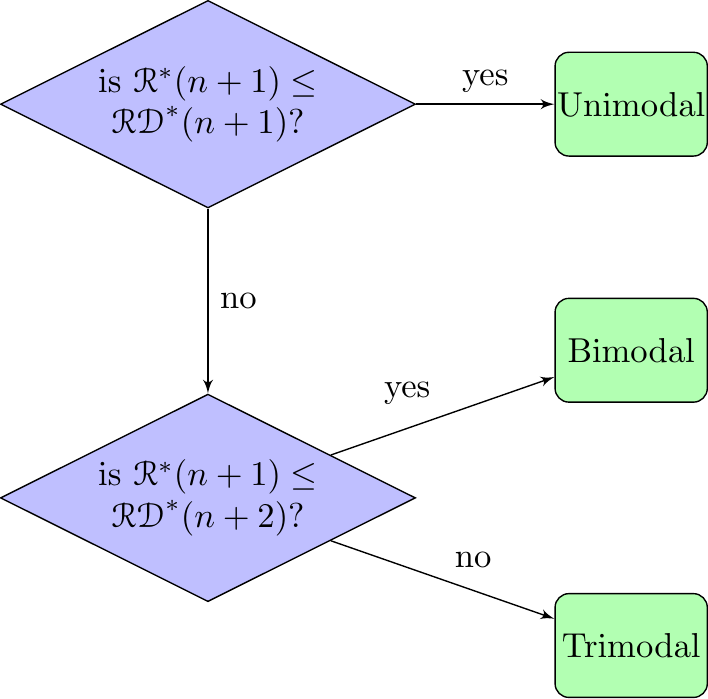}
    \caption{Flowchart summarizing the asymptotic nature of $\mathcal{RD}^*(k)$ based on the relative values of
    $\mathcal{R}^*(n+1)$, $\mathcal{RD}^*(n+1)$ and $\mathcal{RD}^*(n+2)$.}
     \label{fig:flowchart}
\end{figure}

From the plots, it was observed that \emph{beyond $n^2+n+1$ visits, the optimal revisit time is asymptotically periodic with a period $n$.} That is, for $k \geq n^2+n+1$, we have $\mathcal{RD}^*(k+n) = \mathcal{RD}^*(k)$. Furthermore, based on the relative values of $\mathcal{R}^*(n+1)$, $\mathcal{RD}^*(n+1)$ and $\mathcal{RD}^*(n+2)$, the asymptotic behavior can be classified into 3 different types: 1) unimodal; 2) bimodal; and 3) trimodal, as shown in Figure \ref{fig:flowchart}

\subsubsection{Unimodal} When $\mathcal{R}^*(n+1) \leq \mathcal{RD}^*(n+1)$, $\mathcal{RD}^*(k)$ was observed to take exactly one value for all values of $k$ ($\geq n^2+n+1)$), and this value is $\mathcal{RD}^*(n+1)$. Figure \ref{fig:6-4-ort-d} shows the unimodal behavior of $\mathcal{R}^*(k)$.

\subsubsection{Bimodal} When $\mathcal{R}^*(n+1) \leq \mathcal{RD}^*(n+2)$, $\mathcal{RD}^*(k)$ was observed to take at most two distinct values (for $k \geq n^2+n+1$). One of these values is $\mathcal{RD}^*(n+1)$, which is attained when $q = 0$ ($k = pn+1)$. Figure \ref{fig:6-1-ort-d} shows the bimodal behavior of $\mathcal{R}^*(k)$.

\subsubsection{Trimodal} When $\mathcal{R}^*(n+1) > \mathcal{RD}^*(n+2)$, $\mathcal{RD}^*(k)$ was observed to take at most three distinct values (for $k \geq n^2+n+1$). When $q = 1$ ($k = pn+2$), it takes the value $\mathcal{RD}^*(n+2)$, and when $q = 0$ ($k = pn+1$), it takes the value $\mathcal{RD}^*(n+1)$. Figure \ref{fig:6-21-ort-d} shows the trimodal behavior of $\mathcal{R}^*(k)$.

When $k \geq n^2+n+1$, the above-mentioned observations for $q = 0$ and $q = 1$ corroborate Theorems \ref{thm:O1-optimal-q0} and \ref{thm:O2-optimal-q1}. For $q \geq 2$, we tabulate the optimal revisit times alongside their proposed upper bounds (revisit times of the proposed feasible solutions) and lower bounds (from Theorem \ref{thm:lb}) for 40 sample instances in Table \ref{tab:opt_sol}. From the table, it can be seen that the upper bounds match the optimal revisit times in 100 \% of the instances; the same trend was observed in all the 100 instances for which optimal solutions are available. This implies that the \emph{proposed solutions for all these instances are indeed optimal.} Moreover, for 99 of these instances, even the lower bounds match the optimal values, showing the tightness of the proposed bounds. In one instance however, there is a gap of 0.69 \% between the optimal value and the proposed lower bound. This suggests that there is a special case for which the bound can further be tightened.

\begin{table}[h]

    \centering
    \caption{{Quality of proposed solutions and the lower bounds when $q \geq 2$ for sample instances to which optimal solutions are available}}
    \resizebox{0.5\textwidth}{!}{
    \begin{tabular}{|c|c|c|c|c|c|}
        \toprule
        \shortstack{Instance\\ No.} & $n+1$ & \shortstack{$\mathcal{RD}^*(k)$ \\ (for $q \geq 2$)} & U.B. & L.B. & \shortstack{\% gap \\between \\U.B \& L.B.}\\
        \midrule
        1  & 7 & 50.16 & 50.16 & 50.16 & 0 \\
        2  & 7 & 50.93 & 50.93 & 50.93 & 0 \\
        3  & 7 & 51.56 & 51.56 & 51.56 & 0 \\
        4  & 7 & 46.39 & 46.39 & 46.39 & 0 \\
        5  & 7 & 46.28 & 46.28 & 46.28 & 0 \\
        6  & 7 & 41.94 & 41.94 & 41.94 & 0 \\
        7  & 7 & 52.29 & 52.29 & 52.29 & 0 \\
        8  & 7 & 50.91 & 50.91 & 50.91 & 0 \\
        9  & 7 & 46.64 & 46.64 & 46.64 & 0 \\
        10 & 7 & 41.04 & 41.04 & 41.04 & 0 \\
        \midrule
        11 & 6 & 39.22 & 39.22 & 38.98 & 0.62 \\
        12 & 6 & 49.86 & 49.86 & 49.86 & 0    \\
        13 & 6 & 57.90 & 57.90 & 57.90 & 0    \\
        14 & 6 & 50.42 & 50.42 & 50.42 & 0    \\
        15 & 6 & 40.00 & 40.00 & 40.00 & 0    \\
        16 & 6 & 50.20 & 50.20 & 50.20 & 0    \\
        17 & 6 & 42.66 & 42.66 & 42.66 & 0    \\
        18 & 6 & 60.34 & 60.34 & 60.34 & 0    \\
        19 & 6 & 56.70 & 56.70 & 56.70 & 0    \\
        20 & 6 & 44.56 & 44.56 & 44.56 & 0   \\
        \midrule
        21 & 5 & 27.23 & 27.23 & 27.23 & 0 \\
        22 & 5 & 37.85 & 37.85 & 37.85 & 0 \\
        23 & 5 & 49.72 & 49.72 & 49.72 & 0 \\
        24 & 5 & 32.94 & 32.94 & 32.94 & 0 \\
        25 & 5 & 35.07 & 35.07 & 35.07 & 0 \\
        26 & 5 & 42.68 & 42.68 & 42.68 & 0 \\
        27 & 5 & 50.59 & 50.59 & 50.59 & 0 \\
        28 & 5 & 44.6  & 44.6  & 44.6  & 0 \\
        29 & 5 & 33.86 & 33.86 & 33.86 & 0 \\
        30 & 5 & 49.07 & 49.07 & 49.07 & 0 \\
        \midrule
        31 & 4 & 38.07 & 38.07 & 38.07 & 0 \\
        32 & 4 & 34.59 & 34.59 & 34.59 & 0 \\
        33 & 4 & 37.25 & 37.25 & 37.25 & 0 \\
        34 & 4 & 47.91 & 47.91 & 47.91 & 0 \\
        35 & 4 & 30.01 & 30.01 & 30.01 & 0 \\
        36 & 4 & 64.83 & 64.83 & 64.83 & 0 \\
        37 & 4 & 14.05 & 14.05 & 14.05 & 0 \\
        38 & 4 & 39.70 & 39.70 & 39.70 & 0 \\
        39 & 4 & 51.52 & 51.52 & 51.52 & 0 \\
        40 & 4 & 45.50 & 45.50 & 45.50 & 0 \\
        \bottomrule
    \end{tabular}
    }
    \label{tab:opt_sol}
\end{table}

\subsection{Proposed Solutions for Larger Instances}
As mentioned earlier in this section, optimal solutions are not available for larger instances due to the computational difficulty in solving the problem. These instances provide an opportunity to evaluate the usefulness of the proposed heuristics and lower bounds. For 200 instances with $n+1$ ranging from $8$ to $50$, we computed heuristic solutions and the lower bounds, and tabulated the results for sample instances in Table \ref{tab:heuristics}; the table contains results only for the case when $ k \geq n^2+2n+1$ and $q \geq 2$. We skip the discussion on the uninteresting cases of $q = 0$ and $q=1$ for which the results obey Theorems \ref{thm:O1-optimal-q0} and \ref{thm:O2-optimal-q1}.

\begin{table*}[h]
    \centering
    \caption{Table depicting the performance of the proposed solutions and the lower bounds on the optimal revisit times for large instances when $k \geq n^2+2n+1$ and $q \geq 2$}
    \resizebox{\textwidth}{!}{
    \begin{tabular}{|c|c|c|c|c|c|c|c|c|c|c|c|c|}
        \toprule
        \shortstack{Instance\\ No.} & $n+1$ & $\mathcal{R}^*(n+1)$& $\mathcal{RD}^*(n+1)$ & $\mathcal{RD}^*(n+2)$ & $r_1$ & $r_2$ & $r_3$ & \shortstack{Best\\ Solution}& \shortstack{Avg.\\comp.\\ time}& U.B. & L.B & \shortstack{\% gap \\between \\U.B \& L.B.}\\
        \midrule
        41 & 50 & 291.03 & 290.94 & 291.84 & 291.82 & 292.73 & 291.82 & H1,H3    & 1.80 & 291.82 & 291.03 & 0.27 \\
        42 & 50 & 294.68 & 294.50 & 295.08 & 294.68 & 294.68 & 294.68 & H1,H2,H3 & 1.55 & 294.68 & 294.68 & 0    \\
        43 & 50 & 270.03 & 272.22 & 272.85 & 272.22 & 275.89 & 272.22 & H1,H3    & 1.50 & 272.22 & 272.22 & 0    \\
        44 & 50 & 288.97 & 289.26 & 289.96 & 289.26 & 289.26 & 289.26 & H1,H2,H3 & 1.86 & 289.26 & 289.26 & 0    \\
        45 & 50 & 301.97 & 303.38 & 303.97 & 303.38 & 303.38 & 303.38 & H1,H2,H3 & 1.25 & 303.38 & 303.38 & 0    \\
        46 & 50 & 294.64 & 302.00 & 302.00 & 302.00 & 304.45 & 302.00 & H1,H3    & 1.50 & 302.00 & 302.00 & 0    \\
        47 & 50 & 289.83 & 290.65 & 290.65 & 290.65 & 290.65 & 290.65 & H1,H2,H3 & 1.52 & 290.65 & 290.65 & 0    \\
        48 & 50 & 271.69 & 272.04 & 272.04 & 272.04 & 272.04 & 272.04 & H1,H2,H3 & 1.65 & 272.04 & 272.04 & 0    \\
        49 & 50 & 282.31 & 286.18 & 286.88 & 286.18 & 286.18 & 286.18 & H1,H2,H3 & 1.60 & 286.18 & 286.18 & 0    \\
        50 & 50 & 288.90 & 290.60 & 290.60 & 290.60 & 290.60 & 290.60 & H1,H2,H3 & 2.84 & 290.60 & 290.60 & 0    \\
        \midrule
        51 & 30 & 98.32  & 99.51  & 99.51  & 99.51  & 99.51  & 99.51  & H1,H2,H3 & 0.50 & 99.51  & 99.51  & 0    \\
        52 & 30 & 94.91  & 95.05  & 95.05  & 95.05  & 95.30  & 95.05  & H1,H3    & 0.42 & 95.05  & 95.05  & 0    \\
        53 & 30 & 88.95  & 89.29  & 89.29  & 89.29  & 89.29  & 89.29  & H1,H2,H3 & 0.26 & 89.29  & 89.29  & 0    \\
        54 & 30 & 90.50  & 92.43  & 92.43  & 92.43  & 92.43  & 92.43  & H1,H2,H3 & 0.36 & 92.43  & 92.43  & 0    \\
        55 & 30 & 93.33  & 94.41  & 94.41  & 94.41  & 94.41  & 94.41  & H1,H2,H3 & 0.17 & 94.41  & 94.41  & 0    \\
        56 & 30 & 474.91 & 474.72 & 476.70  & 476.67 & 476.82 & 479.77 & H1      & 0.38 & 476.67 & 474.91 & 0.37 \\
        57 & 30 & 425.17 & 424.22 & 427.27 & 425.17 & 425.17 & 425.17 & H1,H2,H3 & 0.39 & 425.17 & 425.17 & 0    \\
        58 & 30 & 415.62 & 415.94 & 417.80  & 416.12 & 416.12 & 416.22 & H1,H2   & 0.46 & 416.12 & 415.94 & 0.04 \\
        59 & 30 & 445.35 & 443.78 & 445.35 & 445.35 & 445.35 & 445.35 & H1,H2,H3 & 0.40 & 445.35 & 445.35 & 0    \\
        60 & 30 & 428.23 & 433.55 & 435.64 & 433.55 & 433.55 & 433.55 & H1,H2,H3 & 0.37 & 433.55 & 433.55 & 0  \\
        \midrule
        61 & 20 & 80.44 & 81.02 & 81.23 & 81.02 & 81.02 & 81.02 & H1,H2,H3 & 0.25 & 81.02 & 81.02 & 0    \\
        62 & 20 & 79.09 & 78.68 & 79.22 & 79.09 & 79.09 & 79.09 & H1,H2,H3 & 0.17 & 79.09 & 79.09 & 0    \\
        63 & 20 & 73.92 & 73.88 & 75.02 & 74.95 & 74.95 & 74.45 & H3       & 0.20 & 74.45 & 73.92 & 0.73 \\
        64 & 20 & 76.66 & 76.36 & 77.06 & 76.66 & 76.66 & 76.66 & H1,H2,H3 & 0.19 & 76.66 & 76.66 & 0    \\
        65 & 20 & 81.89 & 81.99 & 82.75 & 81.99 & 81.99 & 81.99 & H1,H2,H3 & 0.16 & 81.99 & 81.99 & 0    \\
        66 & 20 & 66.08 & 65.53 & 66.36 & 66.36 & 66.43 & 66.36 & H1,H3    & 0.16 & 66.36 & 66.08 & 0.42 \\
        67 & 20 & 81.56 & 81.32 & 81.92 & 81.56 & 81.56 & 81.56 & H1,H2,H3 & 0.15 & 81.56 & 81.56 & 0    \\
        68 & 20 & 80.39 & 80.39 & 80.39 & 80.39 & 80.39 & 80.39 & H1,H2,H3 & 0.17 & 80.39 & 80.39 & 0    \\
        69 & 20 & 79.13 & 77.95 & 79.13 & 79.13 & 79.13 & 79.13 & H1,H2,H3 & 0.17 & 79.13 & 79.13 & 0    \\
        70 & 20 & 73.93 & 74.08 & 75.16 & 74.08 & 74.08 & 74.08 & H1,H2,H3 & 0.12 & 74.08 & 74.08 & 0   \\
        \midrule
        71 & 8 & 49.15 & 47.82 & 48.93 & N/A   & 49.15 & 49.15 & H2,H3    & 0.17 & 49.15 & 49.15 & 0 \\
        72 & 8 & 45.97 & 47.91 & 49.67 & 47.91 & 47.91 & 47.91 & H1,H2,H3 & 0.07 & 47.91 & 47.91 & 0 \\
        73 & 8 & 50.77 & 50.1  & 50.87 & 50.77 & 50.77 & 50.77 & H1,H2,H3 & 0.10 & 50.77 & 50.77 & 0 \\
        74 & 8 & 55.87 & 55.17 & 57.17 & 55.87 & 55.87 & 55.87 & H1,H2,H3 & 0.07 & 55.87 & 55.87 & 0 \\
        75 & 8 & 62.72 & 64.12 & 64.12 & 64.12 & 64.12 & 64.12 & H1,H2,H3 & 0.07 & 64.12 & 64.12 & 0 \\
        76 & 8 & 48.58 & 52.92 & 54.52 & 52.92 & 52.92 & 52.92 & H1,H2,H3 & 0.08 & 52.92 & 52.92 & 0 \\
        77 & 8 & 48.35 & 48.27 & 48.82 & 48.35 & 48.35 & 48.35 & H1,H2,H3 & 0.06 & 48.35 & 48.35 & 0 \\
        78 & 8 & 48.52 & 47.37 & 48.77 & 48.52 & 48.52 & 48.52 & H1,H2,H3 & 0.09 & 48.52 & 48.52 & 0 \\
        79 & 8 & 48.59 & 54.57 & 56.15 & 54.57 & 54.57 & 54.57 & H1,H2,H3 & 0.08 & 54.57 & 54.57 & 0 \\
        80 & 8 & 58.9  & 59.87 & 63.42 & 59.87 & 59.87 & 59.87 & H1,H2,H3 & 0.06 & 59.87 & 59.87 & 0 \\
        \bottomrule
    \end{tabular}
    }
    \label{tab:heuristics}
\end{table*}

\subsubsection{Quality of the proposed heuristics}
For each instance, Table \ref{tab:heuristics} contains the values of $\mathcal{R}^*(n+1)$, $\mathcal{RD}^*(n+1)$, and $\mathcal{RD}^*(n+2)$, the revisit times of the proposed heuristic solutions: $H1$ (if available), $H2$, and $H3$, the best available upper and lower bounds on the revisit times, the type of heuristic that is best for the instance, and the time required for computing all three heuristic solutions. As it can be seen from the table, for most of the instances, the percentage gap between the upper and lower bounds on the optimal revisit time is 0; this is true for 187 out of the 200 instances. This implies that the proposed solutions are proved to be optimal for these instances. For the remaining 13 instances, even though the gap is non-zero, it is not known whether the proposed solution is sub-optimal or if the lower bound is not tight. Irrespective of the reason for the non-zero gap, the average and maximum percentage gap over all the instances are 0.01 \% and 1.04 \% respectively.
Therefore, for all practical purposes, the proposed solutions are near-optimal. Furthermore, the time required to compute a heuristic solution (averaged over all 3 heuristics) for instances with $n+1$ ranging from $8$ to $50$ is a mere 0.73 seconds on an average, whereas it is extremely difficult, if not impossible, to compute optimal solutions. Henceforth, the proposed algorithms demonstrate to be an efficient way to compute high-quality solutions swiftly and facilitate an online implementation.

\subsubsection{Best Heuristic}
Among the proposed heuristic solutions, the ones of the type $H3$ were the best in terms of the revisit time. Out of the 300 instances, solutions of the type $H3$ had the least revisit times in 290 instances, whereas the ones of the type $H2$ and $H1$ had the least revisit times in 250 and 248 instances respectively. With respect to the average computation time, $H2$ only marginally trumps $H3$. The average computation times for $H1$, $H2$, and $H3$ over the 200 instances were 0.7585, 0.7178 and 0.7180 seconds respectively. Therefore, if one decides to implement only one among the three heuristics, $H3$ is the recommended one.



\subsubsection{Results Backing Conjecture \ref{conj:swad_cost}}
Recall that the solutions proposed for the case $\mathcal{R}^*(n+1) > \mathcal{RD}^*(n+2)$ and $q \geq 2$ are \emph{optimal} if Conjecture \ref{conj:swad_cost} is true. To verify this conjecture, in Table \ref{tab:conjecture}, we list the values of $T(\mathcal{W}^1_{STID})$ and $T(\mathcal{W}^*(n+1))$ for all the relevant instances. It can clearly be seen that $T(\mathcal{W}^1_{STID}) \leq T(\mathcal{W}^*(n+1))$ for all these instances, which supports the claims of Conjecture \ref{conj:swad_cost}.

\begin{table}[h]

    \centering
    \caption{Corroboration of Conjecture 1}
    \resizebox{0.5\textwidth}{!}{
    \begin{tabular}{|c|c|c|c|c|}
        \toprule
        \shortstack{Instance\\ No.} & $n+1$ & $\mathcal{R}^*(n+1)$ & $\mathcal{RD}^*(n+2)$ & $T(\mathcal{W}^1_{STID})$\\
        \midrule
        26 & 5  & 42.68  & 41.81  & 41.81  \\
        35 & 4  & 30.01  & 30.01  & 29.05  \\
        71 & 8  & 49.15  & 48.93  & 47.82  \\
        81 & 50 & 281.19 & 280.03 & 280.03 \\
        82 & 7  & 289.66 & 278.76 & 277.99 \\
        83 & 6  & 42.54  & 40.71  & 40.71  \\
        84 & 6  & 47.77  & 47.14  & 45.72  \\
        85 & 6  & 51.06  & 49.97  & 47.52  \\
        86 & 6  & 38.98  & 38.41  & 38.28  \\
        87 & 6  & 38.98  & 38.85  & 38.27  \\
        88 & 5  & 50.24  & 49.10  & 45.12  \\
        89 & 5  & 45.71  & 45.44  & 43.44  \\
        90 & 4  & 41.64  & 39.67  & 38.98  \\
        91 & 4  & 54.79  & 54.49  & 44.77 \\
        \bottomrule
    \end{tabular}
    }
    \label{tab:conjecture}
\end{table}

\subsection{Illustrative Example}
\begin{figure}
\includegraphics[scale=1]{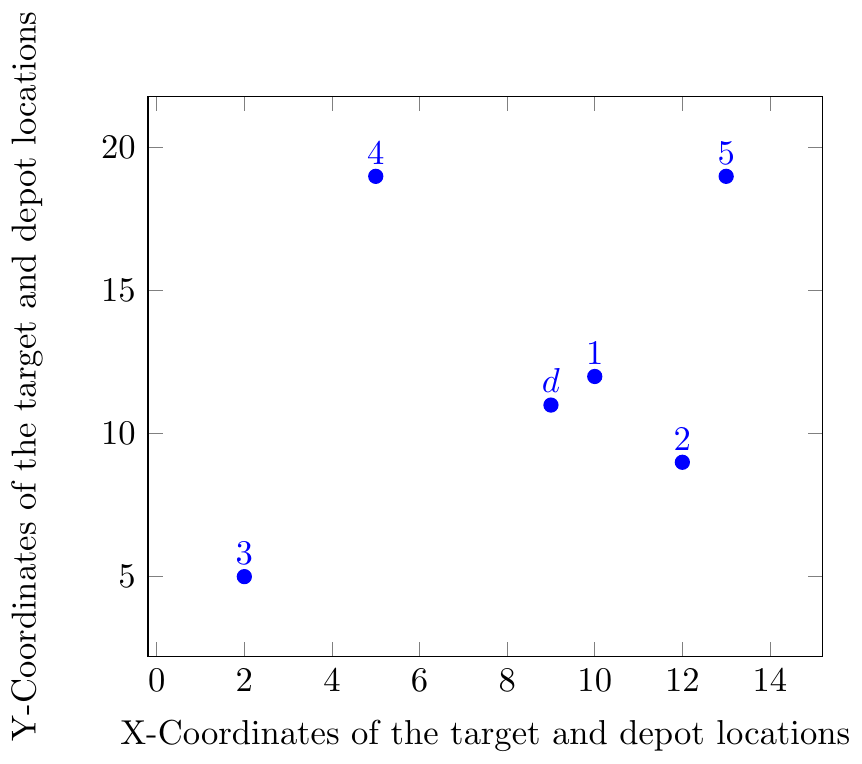}
    \caption{This figures depicts the coordinates of the targets and the depot of a sample 5-target instance.}
    \label{fig:sample-inst-coords}
\end{figure}

Here, we consider a sample instance to illustrate the construction of the proposed solutions for different values of $k$, where $k \geq n^2+2n+1$. The considered instance contains 5 targets and a depot, the locations (coordinates) of which are shown in Figure \ref{fig:sample-inst-coords}.

For this instance, optimal {\prob} walks with $n+1$ and $n+2$ visits, an optimal PMP walk with $n+1$ visits over the $n$ targets, and their revisit times were computed to be the following:


\begin{itemize}

    \item $\mathcal{WD}^*(n+1) =(d, 2, 3, 4, 5, 1, d) $; $\mathcal{RD}^*(n+1) = 45.72 $
    \item $\mathcal{WD}^*(n+2) = (d, 1, 5, 4, 3, 2, 1, d)$ $\mathcal{RD}^*(n+2) = 47.14$
    \item $\mathcal{W}^*(n+1) = (1, 5, 4, 3, 1, 2, 1 )$; $\mathcal{R}^*(n+1) = 47.77$
\end{itemize}

The plot of the optimal revisit times against the number of visits for this instances is shown in Figure \ref{fig:6-21-ort-d}. Clearly, this instance falls under the category $\mathcal{R}^*(n+1) > \mathcal{RD}^*(n+2)$.

When $q = 0$, i.e., for $k = 36, 41, 46,$ etc., an optimal solution is of the type $O1$, and it requires the computation of $\mathcal{WD}^1_{SD}$. This requires us to shortcut the depot from  $\mathcal{WD}^*(n+1)$. To do this, consider a permutation of $\mathcal{WD}^*(n+1)$ about any target, say target 2. The obtained walk is $(2,3,4,5,1,d,2)$. Now, shortcutting the depot from this walk provides $\mathcal{WD}^1_{SD} = (2,3,4,5,1,2)$. The revisit time of this walk was computed to be 44.31 units. Now, suppose the given $k = 36 = 7(5)+1$, we need to concatenate 6 copies of $\mathcal{WD}^1_{SD}$ with a copy of $\mathcal{WD}^*(n+1)$. Since target 2 is visited esxactly once in both the walks, we consider the permutations of these walks about target 2, and concatenate them as the following: $\mathcal{CWD}(36) = (2,3,4,5,1,d,2) \circ \underbrace{(2,3,4,5,1,2) \circ \cdots \circ (2,3,4,5,1,2)}_{\text{6 times}}$. Then, the desired {\prob} walk with 36  visits is obtained by concatenating $\mathcal{CWD}(36)$ about $d$, which gives $\mathcal{WD}^*(36) = (d,2,3,4,5,1,2,3,4,5,1,2, \cdots, 2,3,4,5,1,d)$. It has been verified that the revisit time of this walk is indeed 45.72 units and matches the optimal revisit time.

When $q = 1$, that is, for $k = 37, 42,$ etc., an optimal solution for this case is of the type $O2$. This requires the computation of $\mathcal{WD}^2_{STD}$, which similar to the above discussion, is obtained as $(2,1,5,4,3,2)$, by shortcutting the repeated target 1, and then shortcutting the depot, from $\mathcal{WD}^*(n+2)$. Suppose given $k$ is $37 = 7(5) + 2$, we need to concatenate 6 copies of $\mathcal{WD}^2_{STD}$ with a copy of $\mathcal{WD}^*(n+2)$ as the following: $\mathcal{CWD}(37) = (2,1,d,1,5,4,3,2) \circ \underbrace{(2,1,5,4,3,2) \circ \cdots \circ (2,1,5,4,3,2)}_{\text{6 times}}$. Then, an optimal {\prob} walk with $37$ visits is obtained by permutation $\mathcal{CWD}(37)$ about $d$ as $\mathcal{WD}^*(37) = (d,1,5,4,3,2,1,5,4,3,2 \cdots, 2,1,5,4,3,2,,1,d)$. This walk was verified to have a revisit time of 47.14, which matches the computed optimal revisit time.

When $q \geq 2$, the proposed solution is to compute solutions of the types $H1$ (if possible), $H2$, and $H3$, and choose the best among them. The intermediate walks required to construct these solutions, and their travel times, were computed to be the following:

\begin{itemize}
    \item $\mathcal{WD}^2_{ST} = (d, 1, 5, 4, 3, 2, d)$; $T(\mathcal{WD}^2_{ST}) = 45.72$ 
    \item $\mathcal{WD}^2_{SD}$ does not exist
    \item $\mathcal{W}^1_{ST} = (1, 5, 4, 3, 2, 1) $; $T(\mathcal{W}^1_{ST}) = 44.31$ 
    \item $\mathcal{W}^1_{STID} = (1, 5, 4, 3, 2, d, 1) $; $T(\mathcal{W}^1_{STID}) = 45.72$ 
    \item $\mathcal{WD}^1_{SD} = (2, 3, 4, 5, 1, 2)$; $T(\mathcal{WD}^1_{SD}) = 44.31$. 
    \item $\mathcal{WD}^1_{SDIT} = (2, 1, 3, 4, 5, 1, 2)$; $T(\mathcal{WD}^1_{SDIT}) = 47.77$. 
\end{itemize}

It is to be noted that in this case, shortcutting the depot alone from $\mathcal{WD}^*(n+2)$ is not possible. To see this, perform a permutation of $\mathcal{WD}^*(n+2)$ about target 2, which provides the walk $(2,1,d,1,5,4,3,2)$. In this walk, the target visited before and after the depot is the same (target 1). Therefore, shortcutting the depot will result in an infeasible walk with the adjacent visits being the same. Therefore, in this case, $\mathcal{WD}^2_{SD}$ does not exist, and \textit{computing a solution of the type $H1$ is not possible}.

{\bf Construction of a $H2$ type walk:} Suppose the given number of visits, $k$, is $38$, i.e., $q = 2$. Then, the construction requires the concatenation of a copy of $\mathcal{W}^1_{STID}$, 2 copies of $\mathcal{W}^*(n+1)$ and 4 copies of $\mathcal{W}^1_{ST}$; this constitutes to a total of $6 + 2(6) + 4(5) = 38$ visits. The permutations of $\mathcal{W}^1_{STID}$,  $\mathcal{W}^*(n+1)$ and $\mathcal{W}^1_{ST}$ about target 2 are $(2,d,1,5,4,3,2)$, $(2,1,5,4,3,1,2)$ and $(2,1,5,4,3,2)$ respectively. Then, concatenating these walks as shown in Equation \eqref{eq:concat-feasible-walk-1} provides
$\mathcal{CWD}(38) = (2,d,1,5,4,3,2) \circ (2,1,5,4,3,2) \circ (2,1,5,4,3,1,2) \circ (2,1,5,4,3,1,2) \circ (2,1,5,4,3,2) \circ (2,1,5,4,3,2) \circ (2,1,5,4,3,2)$. Then, permutating $\mathcal{CWD}(38)$ about the depot provides $\mathcal{WD}(38) = (d,1,5,4,3,2,\cdots,2,1,5,4,3,2,d)$. The revisit time of this walk was computed to be 47.77, which matches the optimal revisit time for 38 visits. Therefore, the solution of type $H2$ is optimal for this instance.

{\bf Construction of a $H3$ type walk:} We consider the case $k = 38$. Here, the construction requires the concatenation of a copy of $\mathcal{WD}^*(n+1)$, 2 copies of $\mathcal{WD}^1_{SDIT}$ and 4 copies of $\mathcal{WD}^1_{SD}$. The permutations of $\mathcal{WD}^*(n+1)$, $\mathcal{WD}^1_{SDIT}$ and $\mathcal{WD}^1_{SD}$ are $(2,3,4,5,1,d,2)$, $(2, 1, 3, 4, 5, 1, 2)$, and $(2, 3, 4, 5, 1, 2)$ respectively. Then, concatenating these walks according to Equation \eqref{eq:concat-feasible-walk-1} provides $\mathcal{CWD}(38) = (2,3,4,5,1,d,2) \circ (2, 3, 4, 5, 1, 2) \circ (2, 1, 3, 4, 5, 1, 2) \circ (2, 1, 3, 4, 5, 1, 2) \circ (2, 3, 4, 5, 1, 2) \circ (2, 3, 4, 5, 1, 2) \circ (2, 3, 4, 5, 1, 2)$. Permutation $\mathcal{CWD}(38)$ about the depot provides the desired walk, $\mathcal{WD}(38) = (d,2,3,4,5,1,2, \cdots, 2,3,4,5,1,d)$. The revisit time of this walk is 47.77, which matches the value of $\mathcal{RD}^*(38)$. Therefore, for this instance, the solution of type $H3$ is also optimal.


\section{Conclusion}
\label{sec:conclusions}
An optimal UAV routing problem with an objective of minimizing the maximum revisit time of monitoring targets has been considered in this paper. Most of the existing literature does not account for the fuel capacity of the UAV, the location of its service station and the time required to service the UAV. Here, we consider a computationally challenging version of the problem that simultaneously accounts for the dwell times of the UAV at targets, fuel capacity and the servicing times of the UAV. For this problem we provided tight lower bounds that help in evaluating the quality of any feasible routes developed. We also provided a method to efficiently compute feasible routes and these routes were verified to be optimal or near-optimal for several instances. The findings of the paper suggest that to compute optimal or near-optimal walks with large number ($k \geq n^2+2n+1$) of visits, it is sufficient to compute an optimal {\prob} walk with $n+1$ and $n+2$ visits, and an optimal PMP walk with $n+1$ visits, which are relatively very easy to compute.

The proposed methods adequately solve the problem and make way for considering further extensions that increase the problem's applicability in the real world. Some of these extensions include a gradual transition to weighted targets, considering motion-constrained UAVs, and planning routes for persistent monitoring of targets using multiple UAVs. For insights into these extensions, we refer the readers to the following articles \cite{hari2019bounding, hari2019generalized} and the dissertation \cite{hari2019dissertation}.
\appendices

\section{Proof of Part (ii) of Lemma \ref{lemma:one-rsnd}}
\label{app:vd_n2}

\textit{Part (ii) of Lemma \ref{lemma:one-rsnd}:}
Every {\prob} walk in $\mathcal{M}(k)$ has an r.s.n.d.

\IEEEproof
Recall that $\mathcal{M}$ represents the set of all {\prob} walks with $k \geq n+3$ and $v_d \leq n+2$. We prove the lemma separately for $v_d = n+1$ and $v_d = n+2$.
\subsubsection{$v_d = n+1$}
Let $S$ be a largest r.s.d of $w$ and $t_1$ be its terminus. Then, $S$ has a total of $n+1$ visits; every target is visited exactly once and the remaining visit is to the depot. Without loss of generality, let $S  = \{t_1, t_2, \dots, t_{r-1}, d, t_{r},t_{r+1}, \dots,t_n, t_1\}$. Then, the visit in $w$ immediately following $S$ must be $t_2$. Otherwise, we obtain an r.s.d with $t_2$ as the terminus (starting from the visit to $t_2$ in $S$) that has more than $n+1$ visits. Similarly, the next visit in $w$ must be $t_3$ and so on until $t_{r-1}$.  As every target is visited at least twice, there must be a visit to $t_r$ in $w$ that follows (not necessarily the immediate visit) the visit to $t_{r-1}$. Then, we have a revisit sequence $\{t_r,t_{r+1},\dots t_n, t_1, t_2, \dots, t_{r-1},\dots, t_r\}$ that starts from $t_r$ (the visit in $S$ immediately following $d$), visits every target at least once, and ends at $t_r$ (the visit in $w$). This is the desired r.s.n.d.

\subsubsection{$v_d = n+2$}
In this case, one target is visited twice in the largest r.s.d and the rest of the targets are visited exactly once. Let $S$ be a largest r.s.d and $t_1$ be its terminus. Let the target visited twice in $S$ be $t_r$. Then, $S$ can take two different forms: 1) both the visits to $t_r$ in $S$ are on the same side of the depot, i.e., $S = \{t_1,  \dots, d,\dots, t_r,  \dots  t_r,\dots, t_1\}$ ; or 2) the visits to $t_r$ are on the opposite sides of the depot, i.e.,  $S = \{t_1, \dots, t_r, \dots, d, \dots  t_r, \dots, t_1\}$.

First, let us consider the case in which both the visits to $t_r$ are on the same side of the depot. That is, $S =  \{t_1, t_2,  \dots,t_s, d,t_{s+1}, \dots, t_r,  \dots  t_r,\dots, t_1\}$. Then, the visit in $w$ immediately following $S$ must be $t_2$; otherwise, we obtain an r.s.d with $t_2$ as the terminus that has more than $n+2$ visits. Similarly, all the visits in $S$ prior to the visit to $d$  (i.e., all the visits in $S$ from $t_2$ to $t_s$) must be repeated in the same sequence. Next, let $t_{s+1}$ be the visit immediately following the visit to $d$ in $S$ (if $d$ is immediately followed by $t_r$, let the visit following $t_r$ be referred to as $t_{s+1}$). Because every target is visited at least twice in $w$, $t_{s+1}$ must be visited after (not necessarily immediately) the visit to $t_s$ in $w$. Then, the sequence starting from the visit to $t_{s+1}$ in $S$ and ending at the visit to $t_{s+1}$ in $w$ spans all the targets and is the desired r.s.n.d.

Next, consider the case in which both the visits to $t_r$ are on the opposite sides of the depot. Without loss of generality, let $S = \{t_1, t_2,\dots, t_{r-1}, t_r,t_{r+1}, \dots, t_{s}, d, t_{s+1},\dots  t_r, \dots, t_1\}$. Then, the visit in $w$ immediately following $S$ must be $t_2$; otherwise, we obtain an r.s.d with $t_2$ as the terminus and more than $n+2$ visits. Similarly, all the visits upto $t_{r-1}$ must be repeated in the same order.

Next, the visits from $t_{r+1}$ to $t_s$ must follow (not necessarily in the same order) with one possible visit to another target in beteen; let this other target be referred to as $t_u$. Suppose $t_u \neq t_{s+1}$, where $t_{s+1}$ is the visit immediately following $d$. As every target is visited at least twice in $w$, $t_{s+1}$ is also visited another time in $w$. Then, the revisit sequence starting from the visit to $t_{s+1}$ in $S$ and ending at the visit to $t_{s+1}$ that follows $S$ spans all the targets and is the desired r.s.n.d. Suppose the $t_u = t_{s+1}$. Then, the sequence with $t_{s+2}$ as the terminus satisfies the properties of the desired r.s.n.d.

Hence, every walk in $\mathcal{M}$ has an r.s.n.d.
\endIEEEproof
\bibliographystyle{IEEEtran}
\bibliography{references.bib}

\end{document}